\documentclass[journal]{IEEEtran}
\usepackage{cite,graphicx,subfigure,threeparttable,booktabs,amsmath,amsfonts,amssymb,bm,float}
\usepackage{url}

\ifCLASSINFOpdf
\else
\fi

\begin{document}

\title{Clearing the Skies: A deep network architecture for single-image rain removal}

\author{Xueyang Fu, Jiabin Huang, Xinghao Ding*, Yinghao Liao and John Paisley
\thanks{X. Fu, J. Huang, X. Ding and Y. Liao are with Fujian Key Laboratory of Sensing and Computing for Smart City, School of Information Science and Engineering, Xiamen University. (*Corresponding to: dxh@xmu.edu.cn)}
\thanks{J. Paisley is with the Department of Electrical Engineering, Columbia University, New York, NY 10027 USA.}
\thanks{This work was supported in part by the National Natural Science Foundation of China under Grants 61571382, 61571005, 81301278, 61172179 and 61103121, in part by the Guangdong Natural Science Foundation under Grant 2015A030313007, in part by the Fundamental Research Funds for the Central Universities under Grants 20720160075, 20720150169 and 20720150093, and in part by the Research Fund for the Doctoral Program of Higher Education under Grant 20120121120043.}
}

\maketitle

\begin{abstract}
We introduce a deep network architecture called DerainNet for removing rain streaks from an image. Based on the deep convolutional neural network (CNN), we directly learn the mapping relationship between rainy and clean image detail layers from data. Because we do not possess the ground truth corresponding to real-world rainy images, we synthesize images with rain for training. In contrast to other common strategies that increase depth or breadth of the network, we use image processing domain knowledge to modify the objective function and improve deraining with a modestly-sized CNN. Specifically, we train our DerainNet on the detail (high-pass) layer rather than in the image domain. Though DerainNet is trained on synthetic data, we find that the learned network translates very effectively to real-world images for testing. Moreover, we augment the CNN framework with image enhancement to improve the visual results. Compared with state-of-the-art single image de-raining methods, our method has improved rain 
removal and much faster computation time after network training.
\end{abstract}


\IEEEpeerreviewmaketitle

\section{Introduction}
\begin{figure}[t!]
\centering
\subfigure[Input rainy image]{
\includegraphics[width = .23\textwidth]{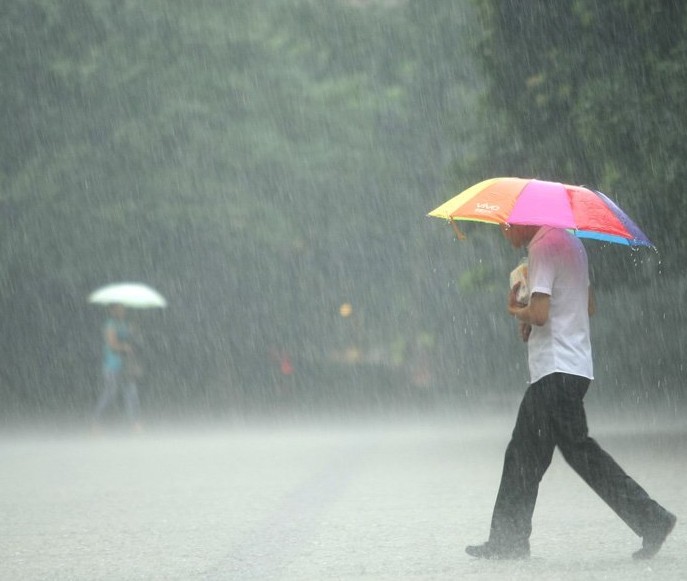}}
\subfigure[Our result]{
\includegraphics[width = .23\textwidth]{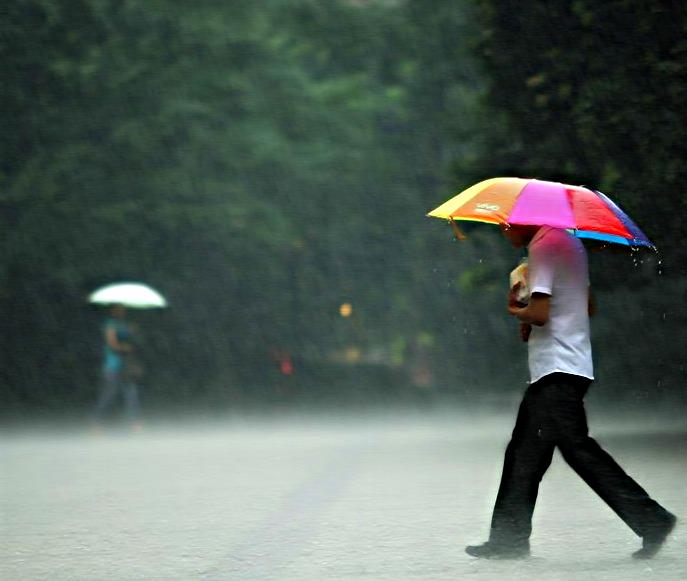}}
\caption{An example real-world rainy image and our de-rained result.} \label{fig.example}
\end{figure}

As the most common bad-weather condition, the effects of rain can degrade the visual quality of images and severely affect the performance of outdoor vision systems. Under rainy conditions, rain streaks create not only a blurring effect in images, but also haziness due to light scattering. Effective methods for removing rain streaks are required for a wide range of practical applications, such as image enhancement and object tracking. We present the first deep convolutional neural network (CNN) tailored to this task and show how the CNN framework can obtain state-of-the-art results. Figure \ref{fig.example} shows an example of a real-world testing image degraded by rain and our de-rained result. 

In the last few decades, many methods have been proposed for removing the effects of rain on image quality. These methods can be categorized into two groups: video-based methods and single-image based methods. We briefly review these approaches to rain removal, then discuss the contributions of our
proposed DerainNet.

\begin{figure*}[ht!]
\centering
\includegraphics[width = 0.9\textwidth]{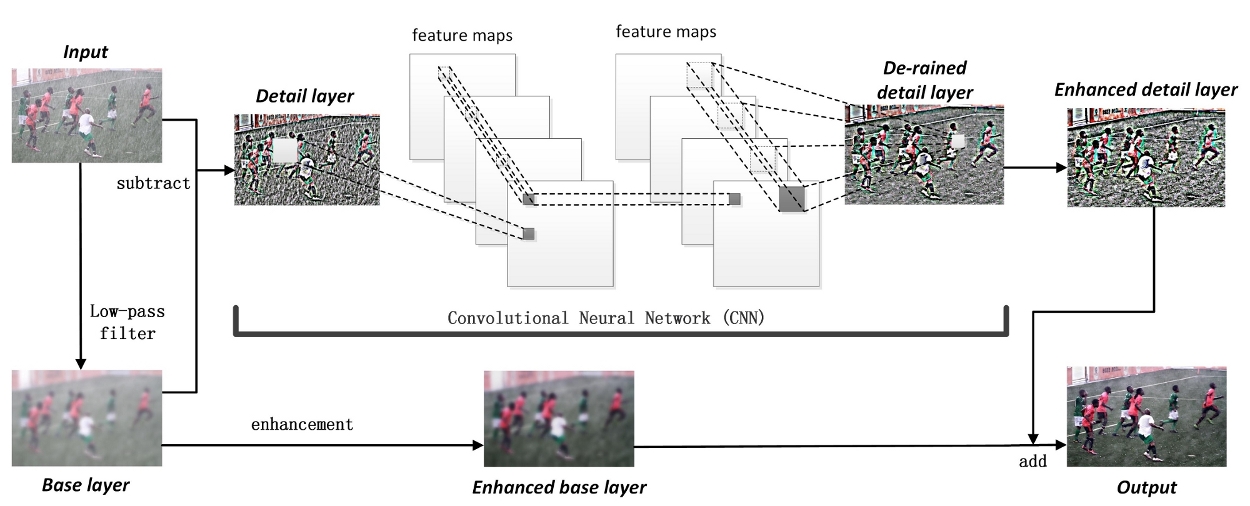}
\caption{The proposed DerainNet framework for single-image rain removal. The intensities of the detail layer images have been amplified for better visualization. } \label{fig.overview}
\end{figure*}

\subsection{Related work: Video v.s.\ single-image based rain removal}
Due to the redundant temporal information that exists in video, rain streaks can be more easily identified and removed in this domain \cite{1,5,6,10}. For example, in \cite{1} the authors first propose a rain streak detection algorithm based on a correlation model. After detecting the location of rain streaks, the method uses the average pixel value taken from the neighboring frames to remove streaks. In \cite{5}, the authors analyze the properties of rain and establish a model of visual effect of rain in frequency space. In \cite{6}, the histogram of streak orientation is used to detect rain and a Gaussian mixture model is used to extract the rain layer. In \cite{10}, based on the minimization of registration error between frames, phase congruency is used to detect and remove the rain streaks. Many of these methods work well, but are significantly aided by the temporal content of video. In this paper we instead focus on removing rain from a single image.

Compared with video-based methods, removing rain from individual images is much more challenging since much less information is available for detecting and removing rain streaks. Single-image based methods have been proposed to deal with this challenging problem, but success is less noticeable than in video-based algorithms, and there is still much room for improvement. To give three examples, in \cite{14} rain streak detection and removal is achieved using kernel regression and a non-local mean filtering. In  \cite{15}, a related work based on deep learning was introduced to remove static raindrops and dirt spots from pictures taken through windows. However, focusing on a specific application this method uses a different physical model from the one in this paper. As our later comparisons show, this physical model limits its ability to transfer to rain streak removal. In  \cite{36}, a generalized low-rank model; both single-image and video rain removal 
can be achieved through this the spatial and temporal correlations learned by this method.

Recently, several methods based on dictionary learning have been proposed  \cite{Huang2012ICME,12,13,16,37}. In \cite{12}, the input rainy image is first decomposed into its base layer and detail layer. Rain streaks and object details are isolated in the detail layer while the structure remains in the base layer. Then sparse coding dictionary learning is used to detect and remove rain streaks from the detail layer. The output is obtained by combining the de-rained detail layer and base layer. A similar decomposition strategy is also adopted in method \cite{37}. In this method, both rain streaks removal and non-rain component restoration is achieved by using a hybrid feature set. In  \cite{13}, a self-learning based image decomposition method is introduced to automatically distinguish rain streaks from the detail layer. In  \cite{16}, the authors use discriminative sparse coding to recover a clean image from a rainy image. A drawback of methods \cite{12,13} is that they tend to generate over-smoothed 
results when dealing with images containing complex structures that are similar to rain streaks, as shown in Figure \ref{fig.synthetic_enlarge}(c), while method \cite{16} usually leaves rain streaks in the de-rained result, as shown in Figure \ref{fig.synthetic_enlarge}(d). Moreover, all four dictionary learning based frameworks \cite{12,13,16,37} require significant computation time. More recently, patch-based priors for both the clean and rain layers have been explored to remove rain streaks \cite{34}. In this method, the multiple orientations and scales of rain streaks are addressed by pre-trained Gaussian mixture models.

\subsection{Contributions of our DerainNet approach}
As mentioned, compared to video-based methods, removing rain from a single image is significantly more difficult. This is because most existing methods \cite{12,13,16,34} only separate rain streaks from object details by using low level features, for example by learning a dictionary to for object representation. When an object's structure and orientation are similar with that of rain streaks, these methods have difficulty simultaneously removing rain streaks and preserving structural information. Humans on the other hand can easily distinguish rain streaks within a single image using high-level features such as context information. We are therefore motivated to design a rain detection and removal algorithm based on the deep convolutional neural network (CNN) \cite{17,lecun1998gradient}. CNN's have achieved success on several low level vision tasks, such as image denoising \cite{18}, super-resolution \cite{19,35}, image deconvolution \cite{20}, image inpainting \cite{22} 
and image filtering \cite{21}. We show that the CNN can also provide excellent performance for single-image rain removal.

In this paper, we propose ``DerainNet'' for removing rain from single-images, which we base on the deep CNN. To our knowledge, this is the first approach based on deep learning to directly address this problem. Our main contributions are threefold:
\begin{enumerate}
 \item DerainNet learns the nonlinear mapping function between clean and rainy detail (i.e., high resolution) layers directly and automatically from data. Both rain removal and image enhancement are performed to improve the visual effect. We show significant improvement over three recent state-of-the-art methods. Additionally, our method has significantly faster testing speed than the competitive approaches, making it more suitable for real-time applications.
 \item  Instead of using common strategies such as increasing neurons or stacking hidden layers to effectively and efficiently approximate the desired mapping function, we use image processing domain knowledge to modify the objective function and improve the de-rain quality. We show how better results can be obtained without introducing more complex network architecture or more computing resources.
 \item  Because we lack access to the ground truth for real-world rainy images, we synthesize a dataset of rainy images using real-world clean images, which we can take as the ground truth. We show that, though we train on synthesized rainy images, the resulting network is very effective when testing on real-world rainy images. In this way, the model can be learned with easy access to an unlimited amount of training data.
\end{enumerate}

\begin{figure*}[th!]
\centering
\includegraphics[width=1.55in]{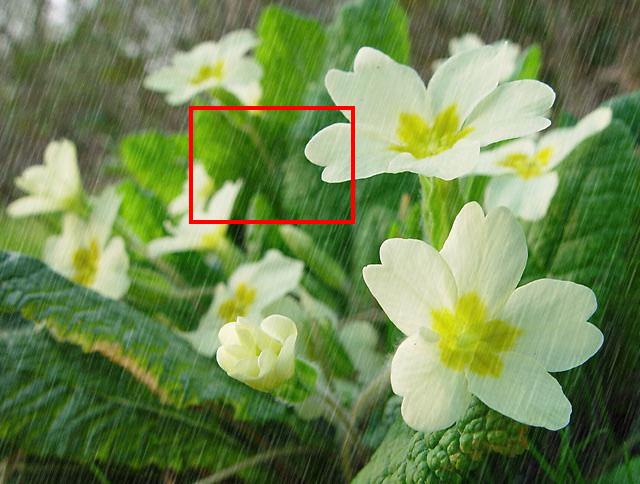}
\includegraphics[width=1.55in]{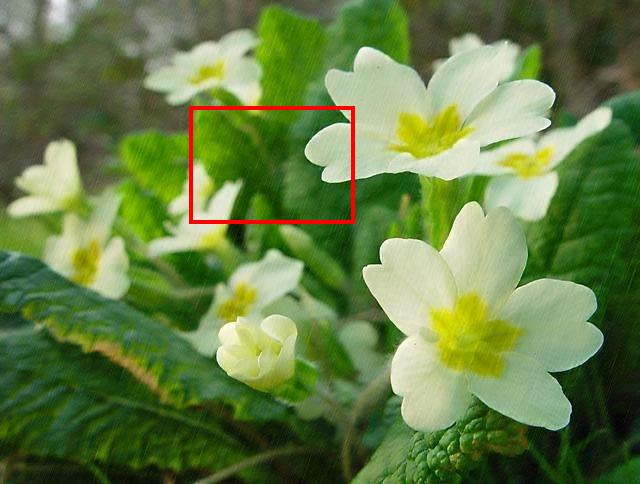}
\includegraphics[width=1.55in]{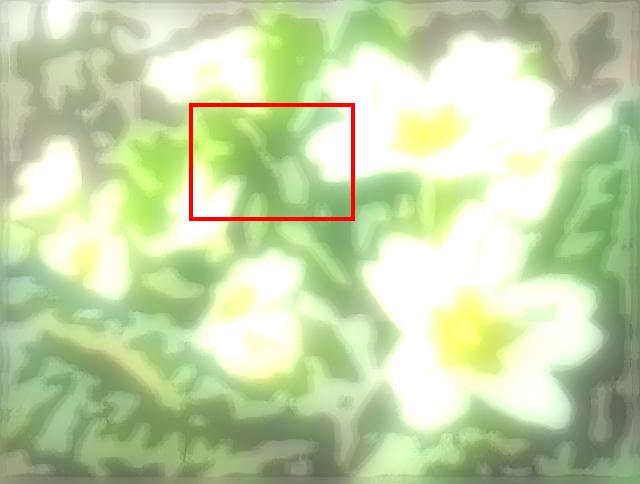}
\includegraphics[width=1.55in ]{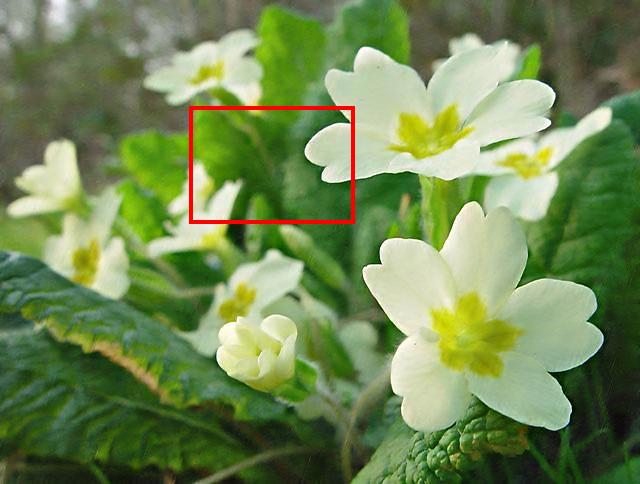}\\
\subfigure[Rainy image]{\includegraphics[width=1.55in ]{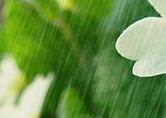}}
\subfigure[Image domain, depth = 3]{\includegraphics[width=1.55in ]{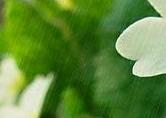}}
\subfigure[Image domain, depth = 10]{\includegraphics[width=1.55in ]{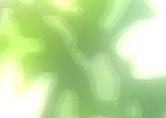}}
\subfigure[Detail layer domain, depth = 3]{\includegraphics[width=1.55in ]{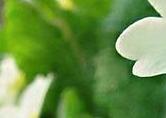}}
\caption{CNN learning options: (b) directly on image domain with depth = 3 (equivalent to retraining \cite{15} on new data), (c) directly on image domain with depth = 10, and (d) on high-frequency detail layer with depth = 3. The first row shows the full image and the second row a zoomed-in region.} \label{fig.domain}
\end{figure*}

\section{DerainNet: Deep learning for rain removal}
We illustrate the proposed DerainNet framework in Figure \ref{fig.overview}. As discussed in more detail below, we decompose each image into a low-frequency base layer and a high-frequency detail layer. The detail layer is the input to the CNN for rain removal. To further improve visual quality, we introduce an image enhancement step to sharpen the results of both layers since the effects of heavy rain naturally leads to a hazy effect.

\subsection{Training on high-pass detail layers}
\label{sec.Detail_strategy}
We denote the input rainy image and corresponding clean image as  $\bf{I}$ and $\bf{J}$ respectively. Initially, a goal may be to train a network architecture ${h_{\bf{P}}}( \cdot )$ that minimizes
\begin{align}
\label{eq.Objective1}
L = \frac{1}{N}\sum_{n=1}^N \| {f_{\bf{W}}}({\bf{I}}^n) - {\bf{J}}^n \|_F^2,
\end{align}
where  ${\bf{W}}$ are the network parameters and $F$ is the Frobenius norm and $n$ indexes the image. However, we found that the result obtained by directly training in the image domain is not satisfactory. In Figure \ref{fig.domain}(a), we show an example of a synthetic rainy image. Note that this image is used in the training process. In Figure \ref{fig.domain}(b) we see that even when this image is used as a training sample, the de-rained image still exhibits clear rain streaks when zoomed in.

Figure \ref{fig.domain}(b) implies that the desired mapping function was not learned well when training on the image domain, i.e., the model under-fit the data. It is natural to ask whether it is necessary to train a more complex model to further improve the capacity of the network. As is well known, there are two ways to improve a network's capacity in the deep learning domain. One way is to increase the depth of network \cite{He2016resnet} by stacking more hidden layers. Usually, more hidden layers can help to obtain high-level features. However, the de-rain problem is a low-level image task and the deeper structure is not necessarily better for this image processing problems. Furthermore, training a feed-forward network with more layers suffers from gradient vanishing unless other training strategies or more complex network structures are introduced. As shown in Figure \ref{fig.domain}(c), when we add network depth to improve the modeling ability, the result actually becomes worse. The other approach is 
to increase the breadth of network  \cite{Schmidhuber} by using more neurons in each hidden layer. However, to avoid over-fitting, this strategy requires more training data and computation time that may be intolerable under normal computing condition.

\begin{figure*}[tp!]
\begin{center}
\includegraphics[height = 0.85in, width = 1.3in]{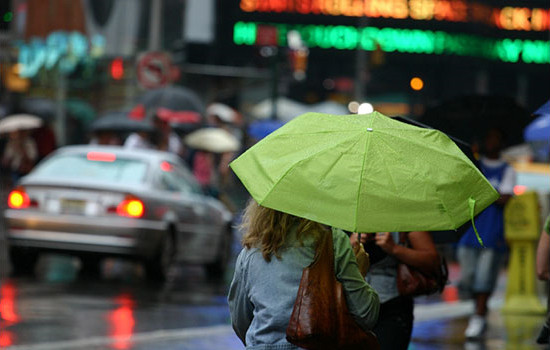}
\includegraphics[height = 0.85in, width = 1.3in]{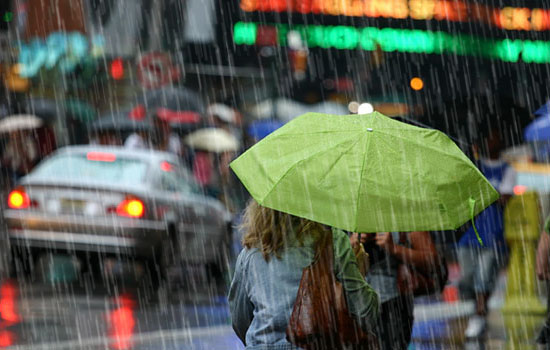}
\includegraphics[height = 0.85in, width = 1.3in]{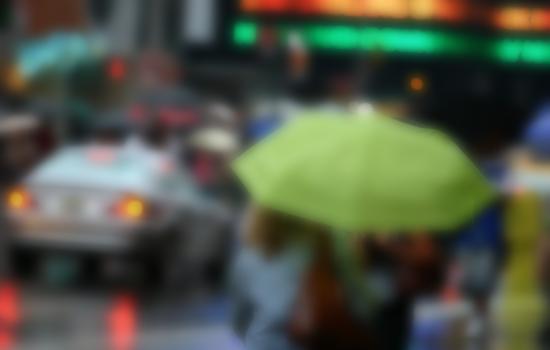}
\includegraphics[height = 0.85in, width = 1.3in]{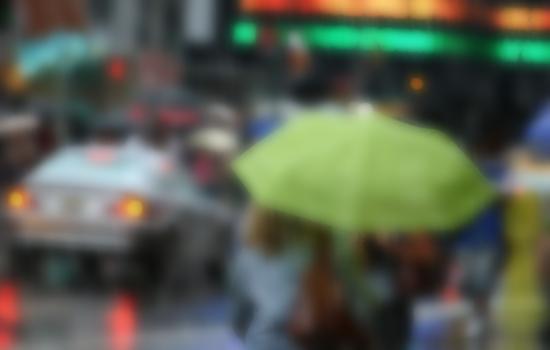}
\includegraphics[height = 0.88in, width = 1.45in]{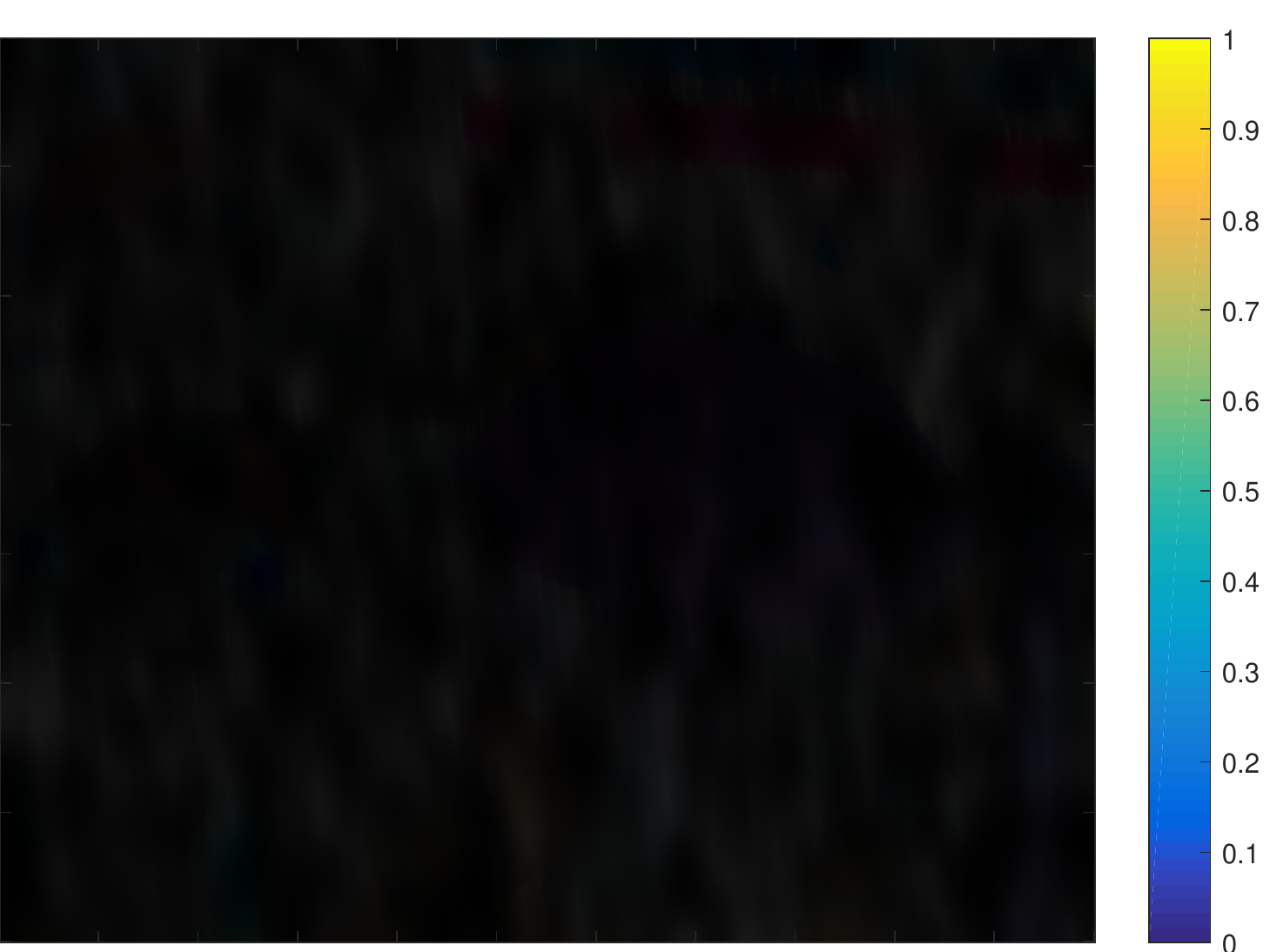}  \\

\subfigure[Clean images ${\bf{J}}$]{\includegraphics[height = 1in, width = 1.3in]{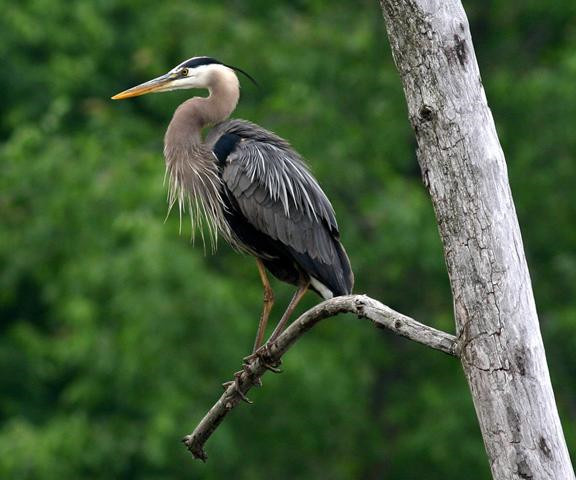}}
\subfigure[Rainy images ${\bf{I}}$]{\includegraphics[height = 1in, width = 1.3in]{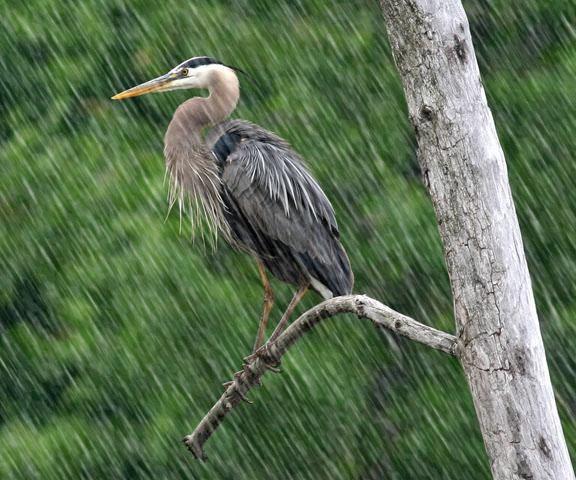}}
\subfigure[${{\bf{J}}_{{\rm{base}}}}$]{\includegraphics[height = 1in, width = 1.3in]{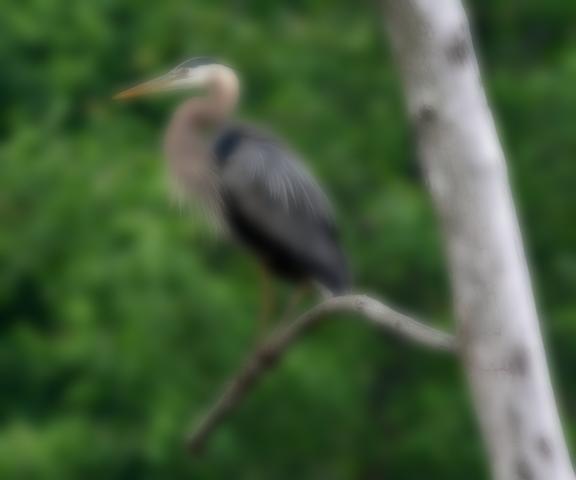}}
\subfigure[${{\bf{I}}_{{\rm{base}}}}$]{\includegraphics[height = 1in, width = 1.3in]{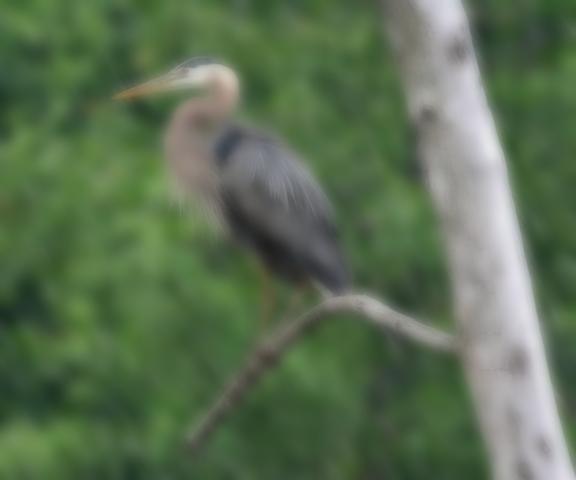}}
\subfigure[$\left| {{{\bf{J}}_{{\rm{base}}}}-{{\bf{I}}_{{\rm{base}}}}} \right|$]{\includegraphics[height = 1.03in, width = 1.45in]{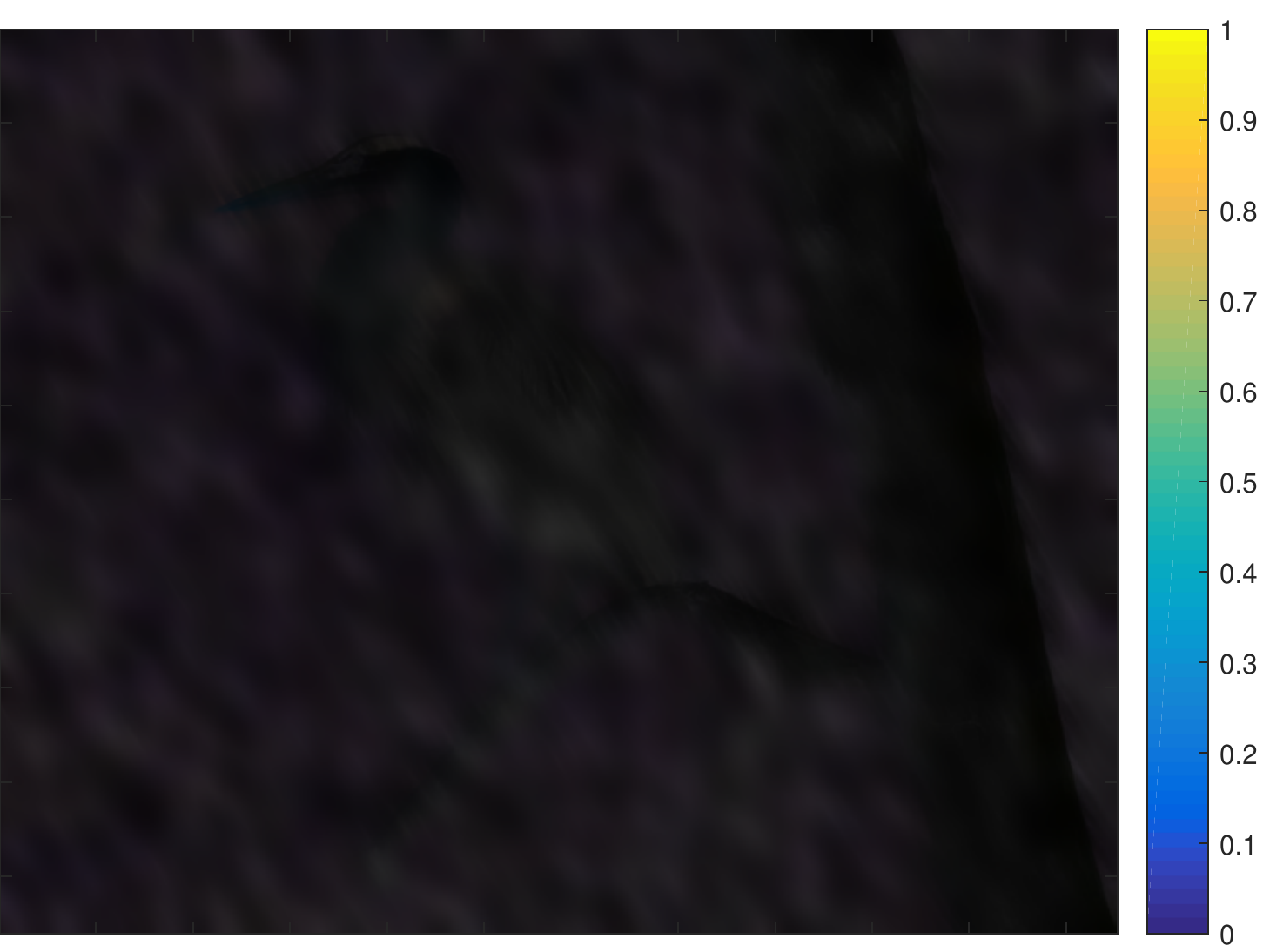}}
\caption{Example base and detail layers of two synthesized images. We use the guided filtering \cite{23} as the low-pass filter to generate the results.} \label{fig.residual}
\end{center}
\end{figure*}

To effectively and efficiently tackle the de-rain problem, we instead use a priori image processing knowledge to modify the objective function rather than increase the complexity of the problem. Conventional end-to-end procedures directly uses image patches to train the model by finding a mapping function $f$ that transforms the input to output \cite{15,19}. Motivated by Figure \ref{fig.domain}, rather than directly train on the image, we first decompose the image into the sum of a ``base'' layer and a ``detail'' layer by using a low-pass filter,
\begin{align}
\label{eq.decompo}
{\bf{J}} = {{\bf{J}}_{{\rm{base}}}} + {{\bf{J}}_{{\rm{detail}}}}.
\end{align}
Using on image processing techniques, we found that after applying an appropriate low-pass filters such as \cite{23,24,25}, low-pass versions of both the rainy image ${{\bf{I}}_{{\rm{base}}}}$ and the clean image ${{\bf{J}}_{{\rm{base}}}}$ are smooth and are approximately equal, as shown in Figure \ref{fig.residual}. In other words, both the rain streaks and the object's details remain in the high-pass detail layer and ${{\bf{I}}_{{\rm{base}}}} \approx {{\bf{J}}_{{\rm{base}}}}$. This implies that the base layer portion can be removed from the training process, significantly simplifying the mapping needed to be learned by the CNN. Thus, we rewrite the objective function in (\ref{eq.Objective1}) as
\begin{align}
\label{eq.spread}
L =  \frac{1}{N}\sum_{n=1}^N\left\| {f_{\bf{W}}}({{\bf{I}}^n_{{\rm{detail}}}}) - {{\bf{J}}^n_{{\rm{detail}}}} \right\|_F^2.
\end{align}

\begin{figure*}[tp!]
\begin{center}
\subfigure[Clean image ${\bf{J}}$]{\includegraphics[width = 1.6in]{results/2domain/1_original.jpg}}
\subfigure[Rainy image ${\bf{I}}$]{\includegraphics[width = 1.6in]{results/2domain/1_rain.jpg}}
\subfigure[${{\bf{J}}_{{\rm{detail}}}}$]{\includegraphics[width = 1.6in]{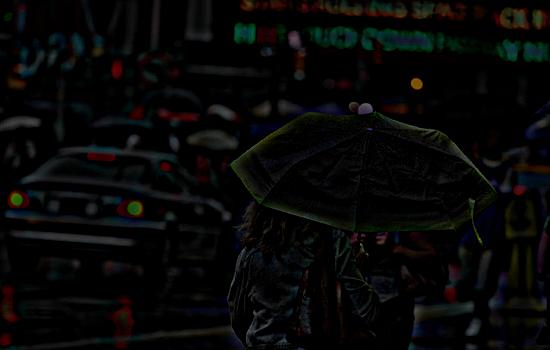}}
\subfigure[${{\bf{I}}_{{\rm{detail}}}}$]{\includegraphics[width = 1.6in]{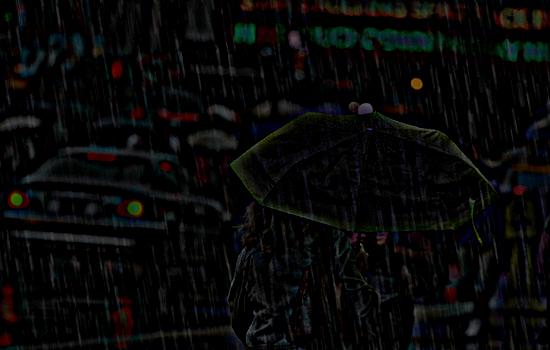}}\\
\subfigure[Histogram of (a)]{\includegraphics[width = 1.6in]{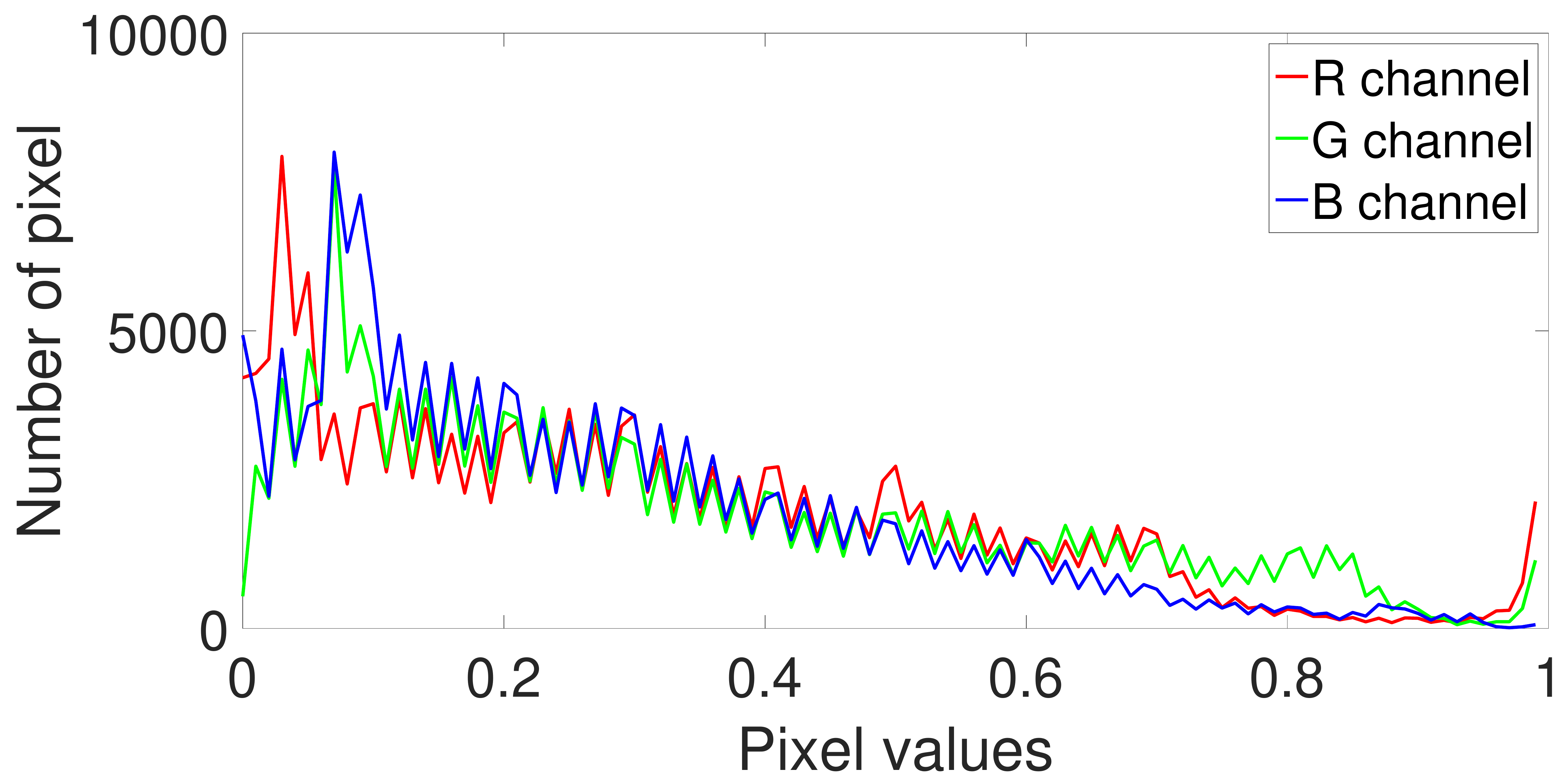}}
\subfigure[Histogram of (b)]{\includegraphics[width = 1.6in]{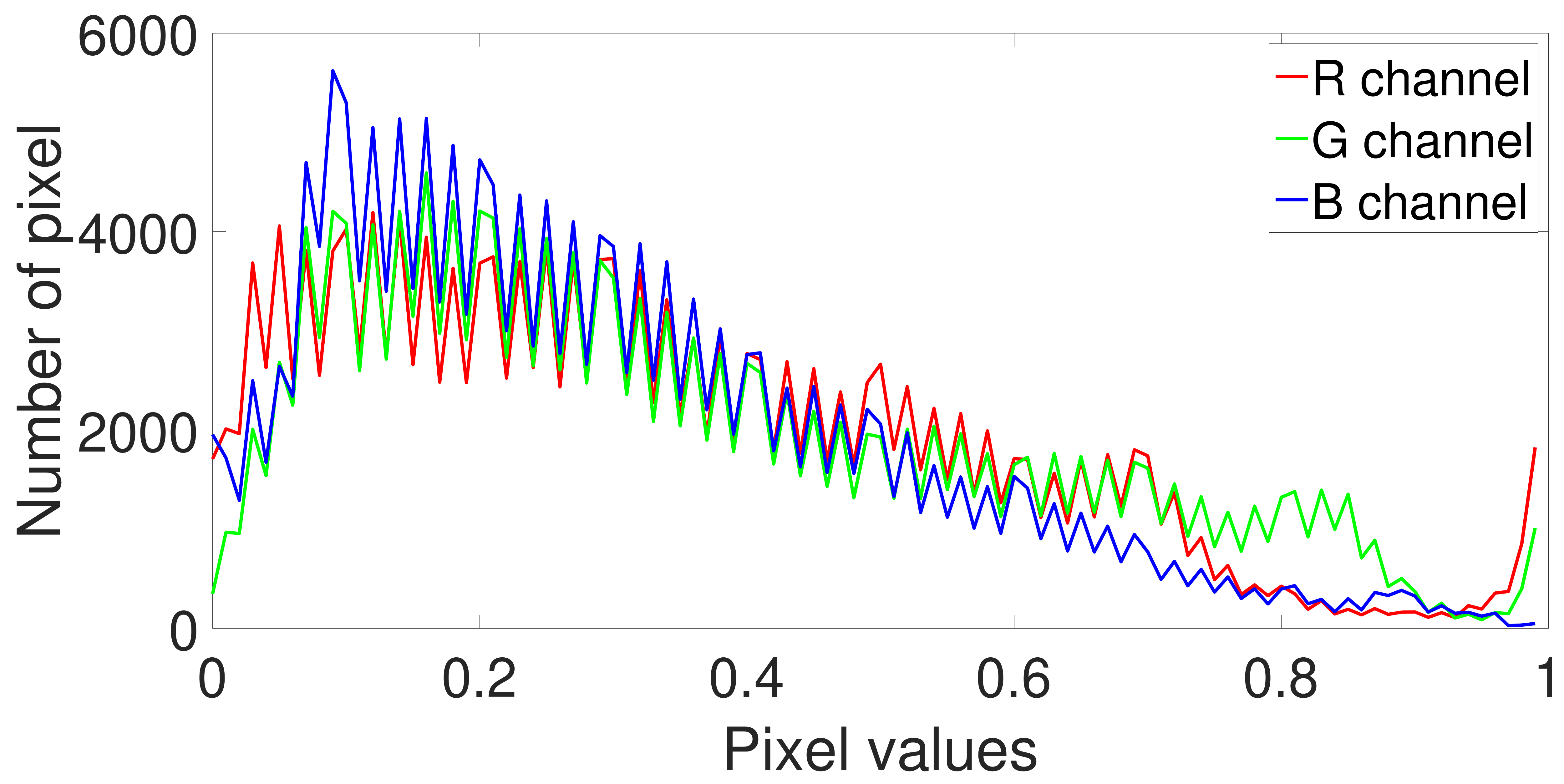}}
\subfigure[Histogram of (c)]{\includegraphics[width = 1.6in]{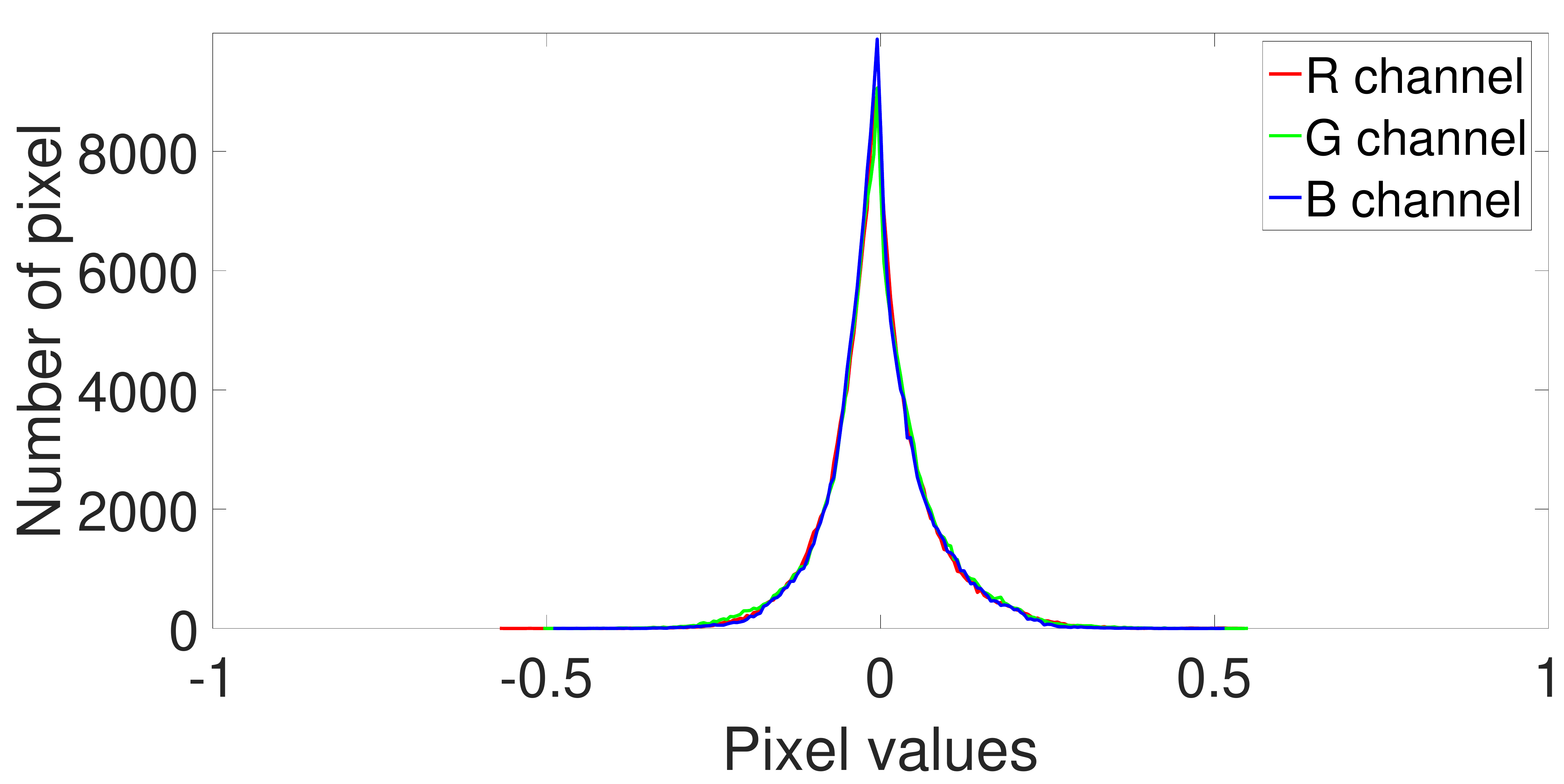}}
\subfigure[Histogram of (d)]{\includegraphics[width = 1.6in]{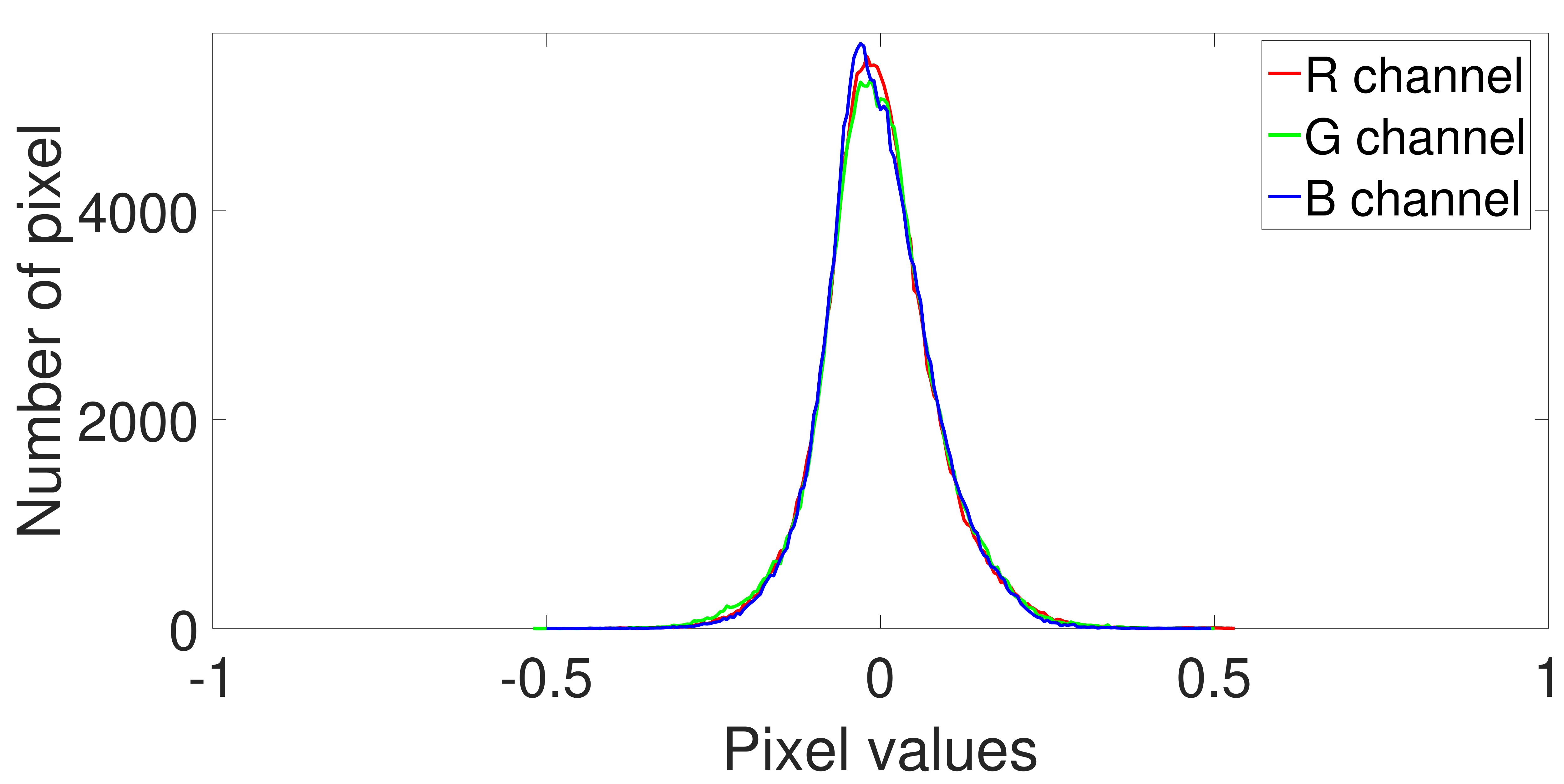}}
\caption{Sparsity of detail layer. The detail layers are obtained by ${{\bf{J}}_{{\rm{detail}}}}={\bf{J}} - {{\bf{J}}_{{\rm{base}}}}$ and ${{\bf{I}}_{{\rm{detail}}}}={\bf{I}} - {{\bf{I}}_{{\rm{base}}}}$.} \label{fig.detail}
\end{center}
\end{figure*}

This directly lead us to train the CNN network on the detail layer instead of the image domain. Moreover, training on the detail layer has several advantages. First, after subtracting the base layer, the detail layer is sparser than the image since most regions in the detail layer are close to zero. As shown in Figure \ref{fig.detail}, the detail layer has many more pixels that are close to zero than the image itself. Taking advantage of the sparsity of the detail layer is a widely used technique in existing de-raining methods \cite{12,13,16}. In the context of a neural network, training a CNN on the detail layer also follows the procedure of mapping an input patch to an output patch, but since the mapping range has been significantly decreased, the regression problem is significantly easier to handle for a deep learning model. Thus, training on the detail layer instead of the image domain can improve learning the network weights and thus the de-raining result without a large increase in training data or 
computational resources.

A second advantage of training on sparse data is that it can improve the convergence of the CNN. As we show in our experiments (Figure \ref{fig.convergence}), training on the detail layer converges much faster than training on the image domain. A third advantage is that decomposing an image into base and detail layers is widely used by the wider image enhancement community \cite{Gu2013HDR,Qiu2013LLSURE}. These enhancement procedures are tailored to this decomposition and can be easily embedded into our architecture to further improve image quality, which we describe in Section \ref{sec.enhancement}.

%
We therefore first decompose the image into a base layer by using a low-pass filter and a detail layer; the detail layer is equal to the difference between the image and the base layer. We use the guided filtering method of \cite{23} as the low-pass filter because it is simple and fast to implement. In this paper, the guidance image is the input image itself.
However, the choice of low-pass filter is not limited to guided filtering; other filtering approaches were also effective in our experiments, such as bilateral filtering \cite{24} and rolling guidance filtering \cite{25}. Results with these filters were nearly identical, so we choose \cite{23} for its low computational complexity. 

After this decomposition we train the CNN on the detail layer image instead of raw image itself according to Eq.\ (\ref{eq.spread}).
This step represents the CNN portion of Figure \ref{fig.overview}. In Figure \ref{fig.domain}(d) we show an example of the de-rained image using this training approach. In terms of rain streak removal, the result is clearly better than the same CNN structure trained on the image domain shown in Figure \ref{fig.domain}(b). This conclusion is further supported by our experiments below.

\begin{figure*}[t]
\centering
\subfigure[Rainy image]{
\includegraphics[height = 1in, width = 1.6in]{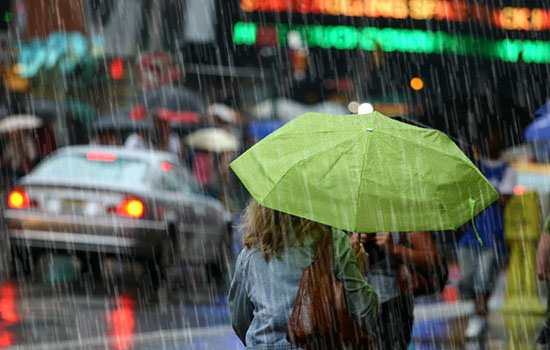}}
\subfigure[Our result]{
\includegraphics[height = 1in, width = 1.6in]{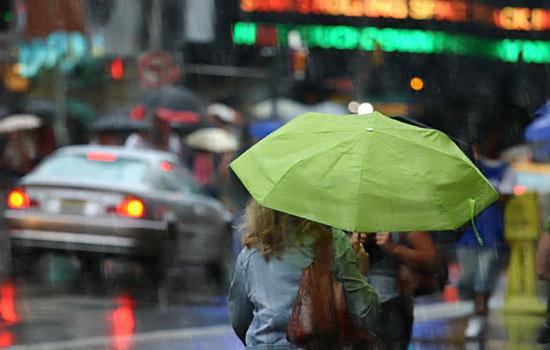}}
\subfigure[${{\bf{W}}^1}$]{
\includegraphics[height = 1in, width = 1.6in]{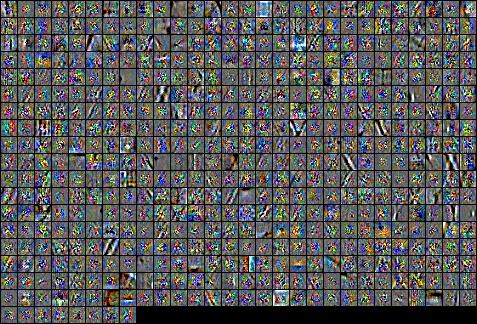}}
\subfigure[${{\bf{W}}^3}$]{
\includegraphics[height = 1in, width = 1.6in]{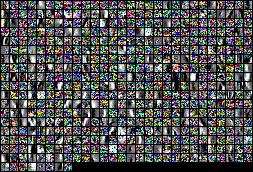}}\\
\subfigure[${\bf{I}_{{\rm{detail}}}}$]{
\includegraphics[height = 1in, width = 1.6in]{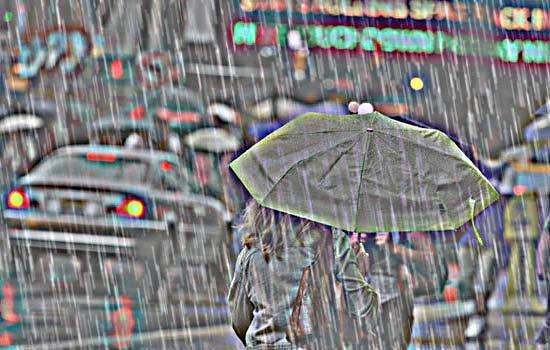}}
\subfigure[${f^1}({\bf{I}_{{\rm{detail}}}})$ ]{
\includegraphics[height = 1in, width = 1.6in]{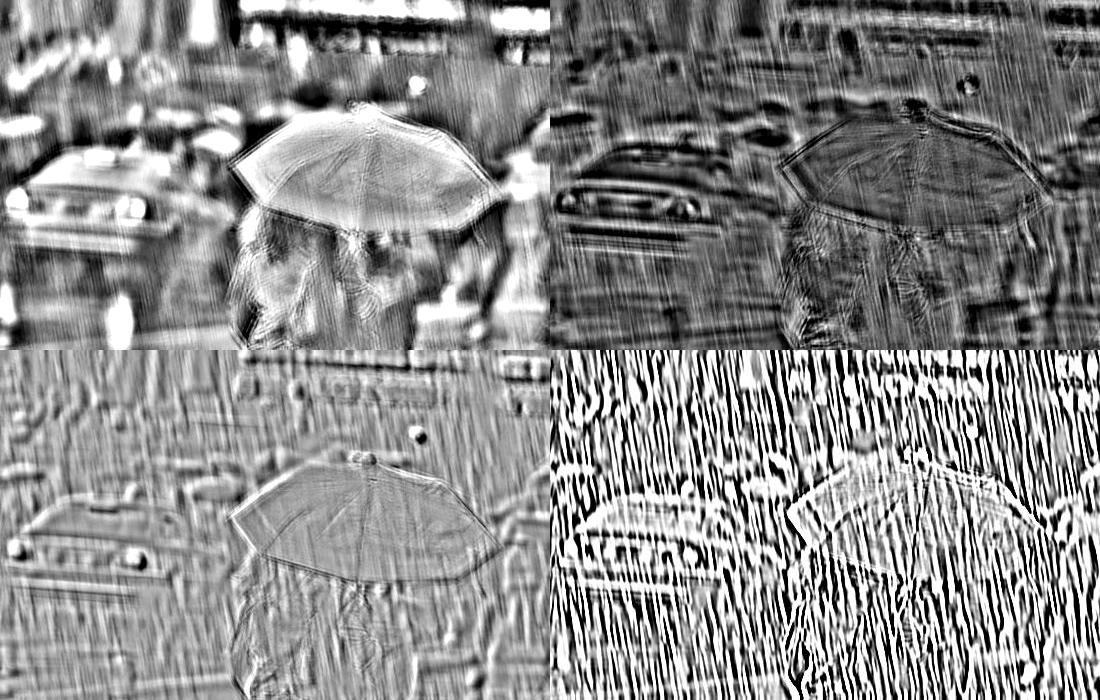}}
\subfigure[${f^2}({\bf{I}_{{\rm{detail}}}})$ ]{
\includegraphics[height = 1in, width = 1.6in]{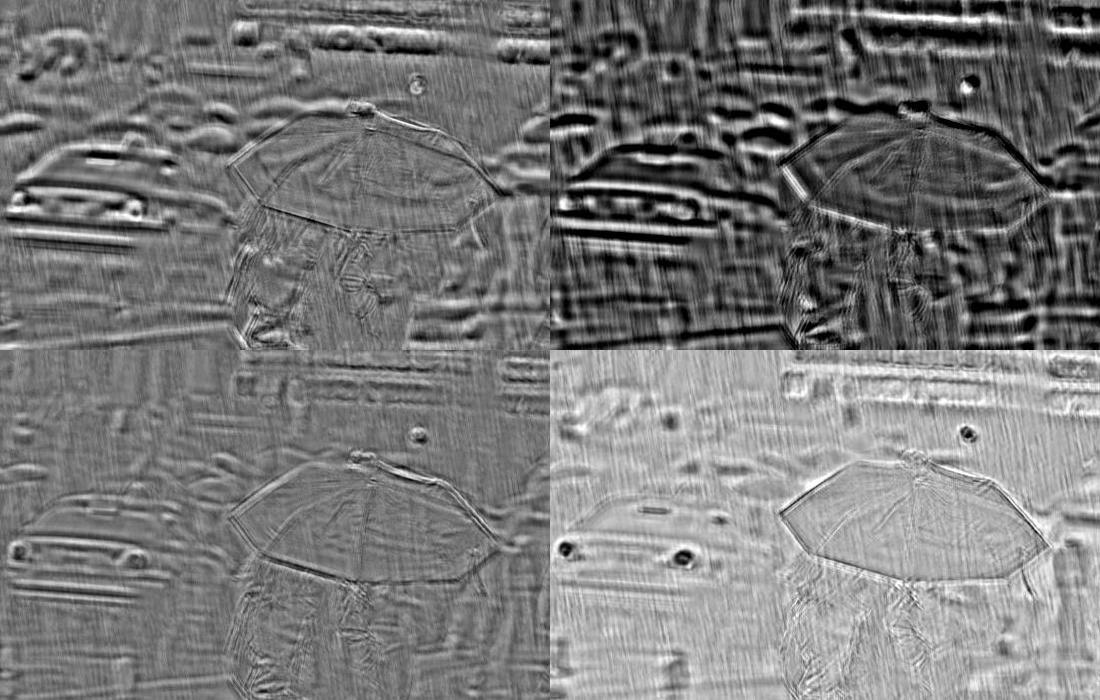}}
\subfigure[De-rained ${f_{\bf{W}}}({\bf{I}_{{\rm{detail}}}})$]{
\includegraphics[height = 1in, width = 1.6in]{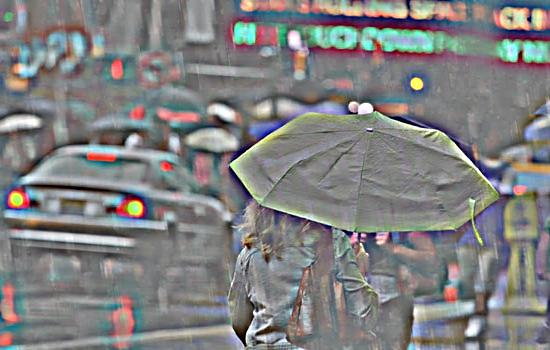}}
\caption{Visualization of intermediate results. The first row shows our de-raining result and the trained weights ${{\bf{W}}^1}$ (512 kernels of size $16\times16\times 3$) and ${{\bf{W}}^3}$ (3 kernels of size $8\times8\times 512$, one for each color channel). For ${{\bf{W}}^3}$ we visualize these three kernels as RGB images across the 512 dimensions.  Since the 512 kernels ${{\bf{W}}^2}$ are $1\times1\times 512$, we do not show them. The second row shows the corresponding hidden layer activations. (e) and (h) show the detail layer input and output of the network. In (f) we show four of the 512 convolutional output of (e) in the first layer. These appear to be producing different ``views'' of the rain. In (g) we show four of the 512 layers that are combined to produced the three layer RGB output in (f).  The intensities of the images in the second row have been amplified for better visualization.} \label{fig.intermediate}
\end{figure*}

\begin{figure*}[th!]
\centering
\includegraphics[width = .78in]{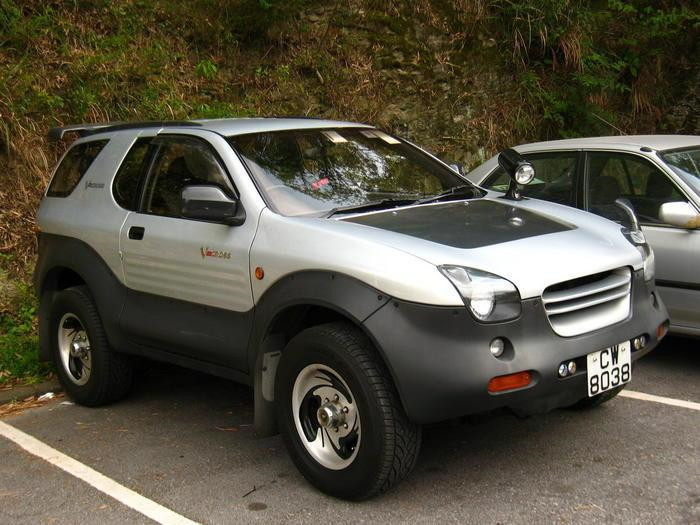}\vspace{1pt}
\includegraphics[width = .78in]{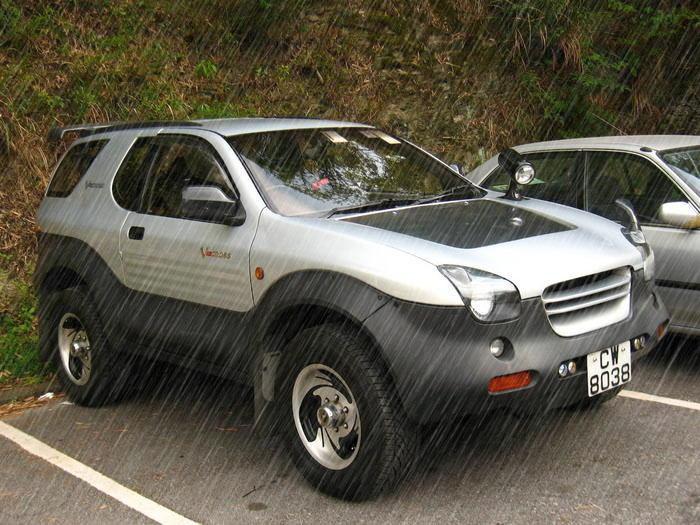}
\includegraphics[width = .78in]{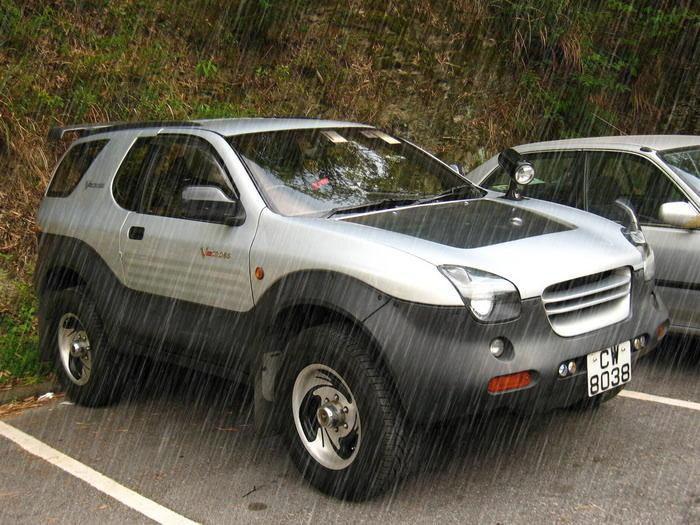}
\includegraphics[width = .78in]{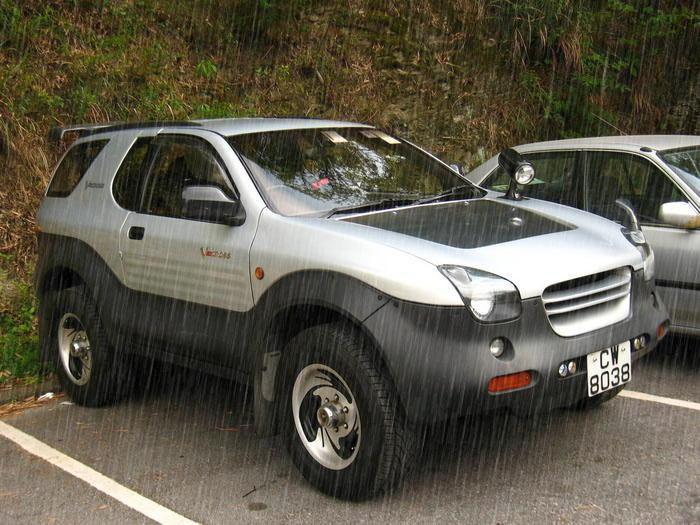}
\includegraphics[width = .78in]{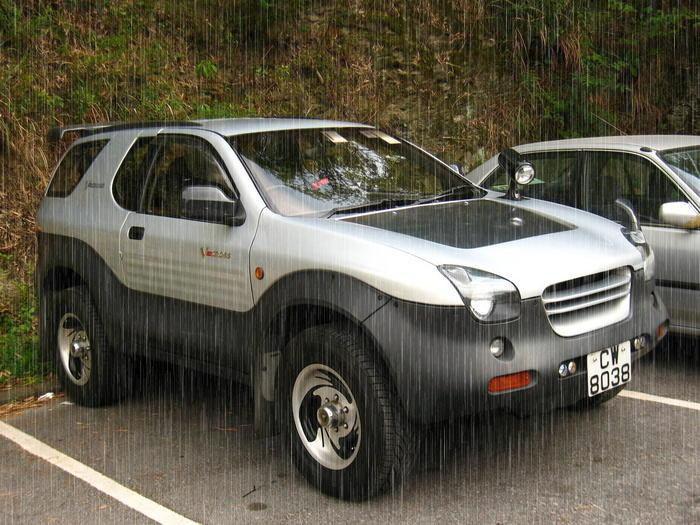}
\includegraphics[width = .78in]{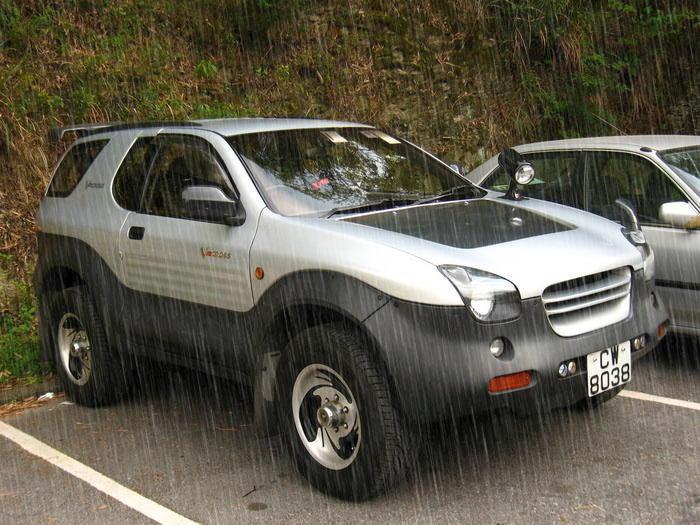}
\includegraphics[width = .78in]{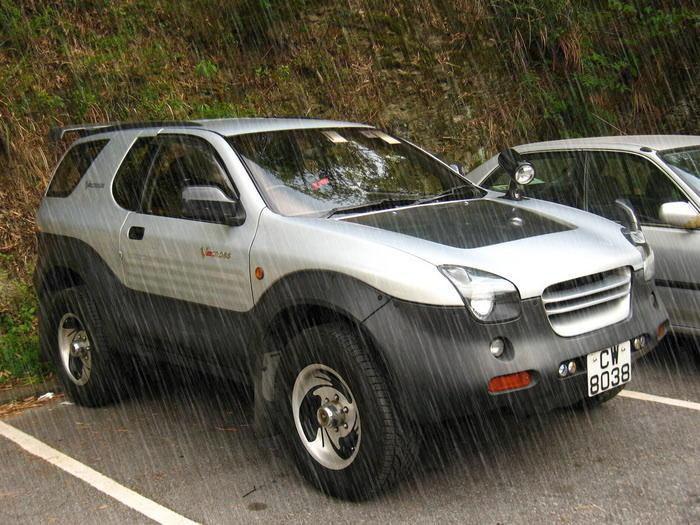}
\includegraphics[width = .78in]{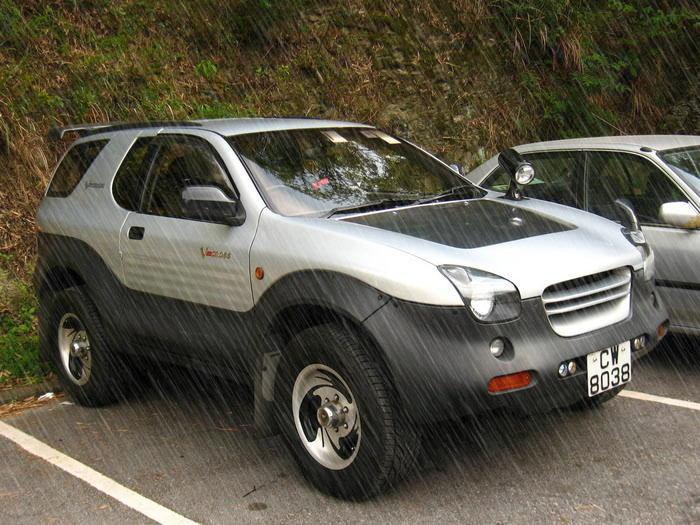}\\
\includegraphics[width = .78in]{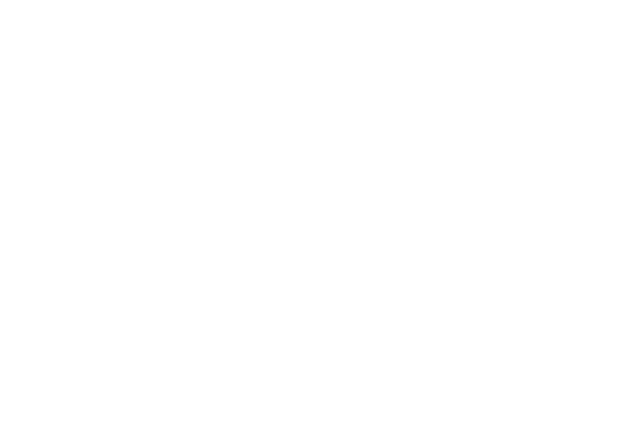}
\includegraphics[width = .78in]{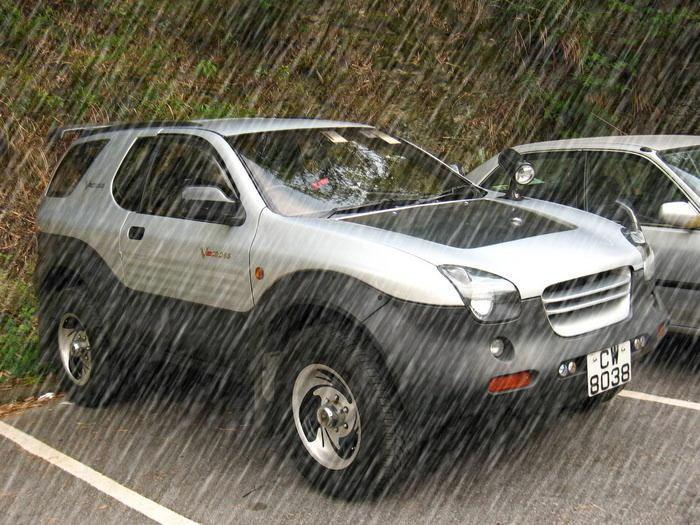}
\includegraphics[width = .78in]{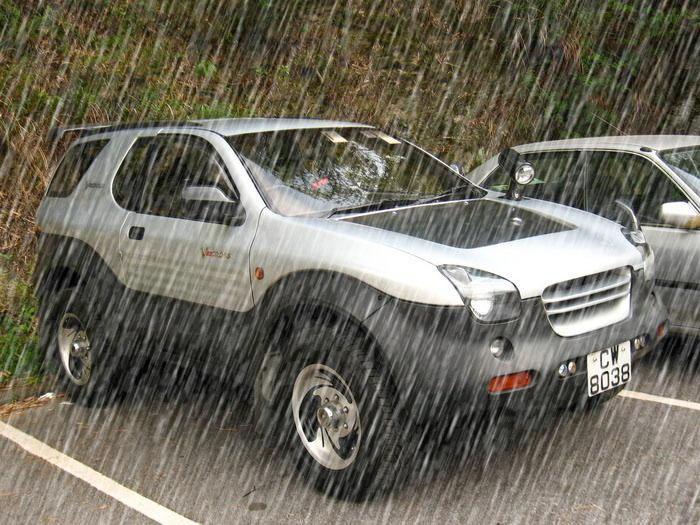}
\includegraphics[width = .78in]{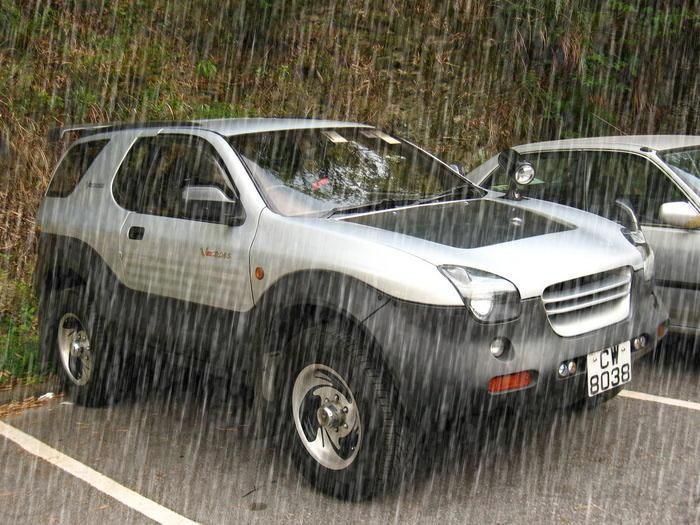}
\includegraphics[width = .78in]{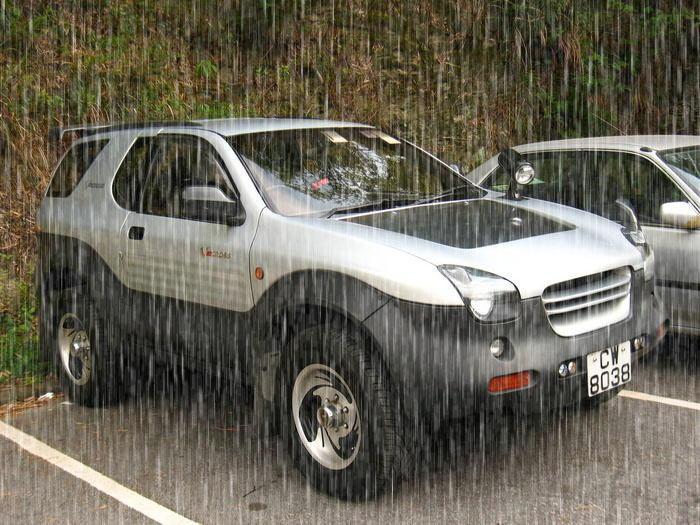}
\includegraphics[width = .78in]{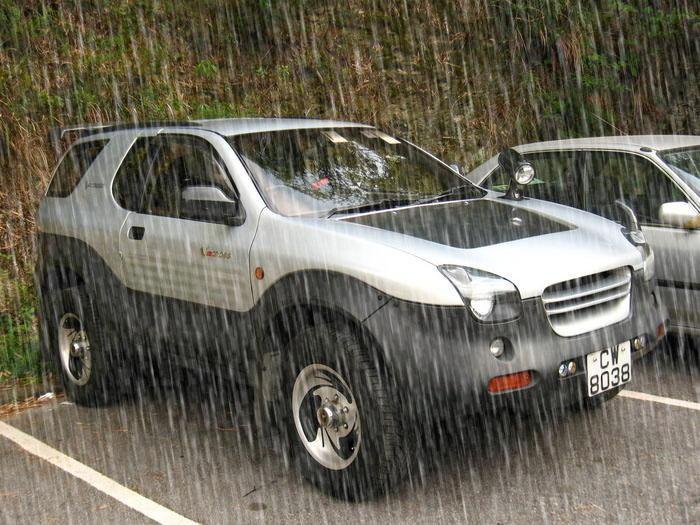}
\includegraphics[width = .78in]{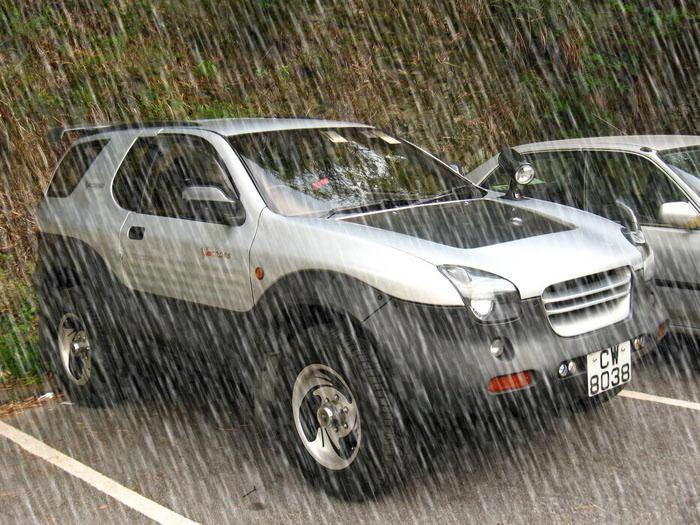}
\includegraphics[width = .78in]{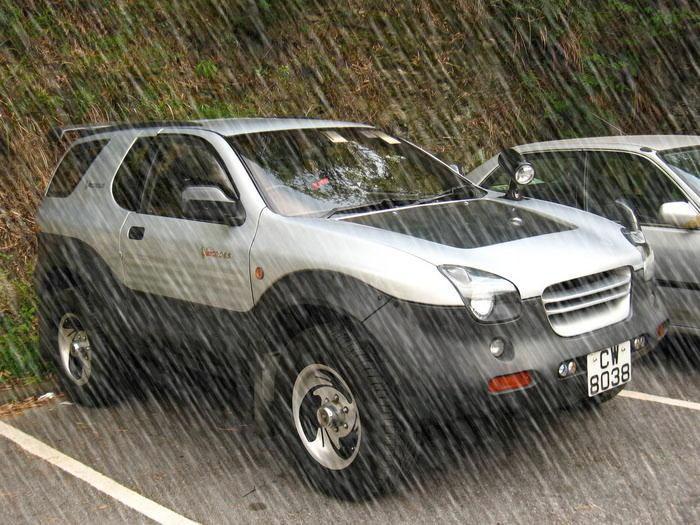}
\caption{An example of synthesized rainy images. The top left is the clean image and the remaining are various images synthesized.} \label{fig.examplesdata}
\end{figure*}

\subsection{Our convolutional neural network}\label{sec.ourCNN}
Our network structure can be expressed as three operations:
\begin{align}
\label{eq.network1} {f^l}({\bf{I}_{{\rm{detail}}}}) &= \sigma ({{\bf{W}}^l}*{f^{l - 1}}({\bf{I}_{{\rm{detail}}}}) + {\bf{b}}^l),\ \ l = 1,2\\
\label{eq.network2} {f_{\bf{W}}}({\bf{I}_{{\rm{detail}}}}) &= {{\bf{W}}^l}*{f^{l - 1}}({\bf{I}_{{\rm{detail}}}}) + {\bf{b}}^l,\ ~~~~~~\, l = 3,
\end{align}
where  $l$ indexes layer number, $*$  indicates the convolution operation and ${b^l}$ is the bias. We define $\sigma ( \cdot )$ to be the nonlinear hyperbolic tangent function and ${f^0}({\bf{I}_{{\rm{detail}}}}) = {\bf{I}_{{\rm{detail}}}}$. We use two hidden layers in our DerainNet architecture and Eq.\ (\ref{eq.network2}) is the output of the cleaned detail layer.


To better understand the effects of the network ${f_{\bf{W}}}$, we show the learned weights and intermediate results from the hidden layers in Figure \ref{fig.intermediate}. The first hidden layer performs feature extraction on the input detail layer, which is similar to the common strategy used for image restoration of extracting and representing image patches by a set of dictionary elements. Thus, ${{\bf{W}}^1}$  contains some filters that look like edge detectors that align with the direction of rain streaks and object edges. The second hidden layer performs the rain streaks removal and ${f^2}({\bf{I}_{{\rm{detail}}}})$  looks smoother than  ${f^1}({\bf{I}_{{\rm{detail}}}})$. The third layer performs reconstruction and enhances the smoothed details with respect to image content. As can be seen in Figure \ref{fig.intermediate}, ${f_{\bf{W}}}({\bf{I}_{{\rm{detail}}}})$ contains clear details with most of the rain removed. The intermediate results show that the CNN is effective at feature extraction and 
helps to recognize and remove rain streaks.

\subsection{Training}
We use stochastic gradient descent (SGD) to minimize the objective function in Eq. (\ref{eq.spread}). Since it is extremely difficult to obtain a large number of clean/rainy image pairs from real-world data, we synthesize rain using Photoshop\footnote{\url{http://www.photoshopessentials.com/photo-effects/rain/}} to create our training dataset. We randomly collected a total of 350 clean outdoor images from the UCID dataset \cite{27}, the BSD dataset \cite{31} and Google image search which we used to synthesize rainy images. Each clean image was used to generate $14$ rainy images of different streak orientations and intensity. An example is shown in Figure \ref{fig.examplesdata}. Thus we create a dataset containing $350 \times 14 = 4900$ rainy images, each having a corresponding ground truth clean image. We randomly selected one million $64 \times 64$ clean/rainy patch pairs from this synthesized data as training samples. A $56 \times 56$ output is generated to avoid border effects caused by convolution. In 
each iteration, $t$, the CNN weight and bias are updated using back-propagation,
\begin{align}
\label{eq.SGD}
{\bf{W}_{t + 1}} &= {\bf{W}_t} - \alpha {({f_{\bf{W}}}({\bf{I}_{{\rm{detail\_i}}}}) - {\bf{J}_{{\rm{detail\_i}}}})^T}\frac{{\partial {f_{\bf{W}}}({\bf{I}_{{\rm{detail\_i}}}})}}{{\partial \bf{W}}}, \nonumber \\
{{\bf{b}}_{t + 1}} &= {{\bf{b}}_t} - \alpha {({f_{\bf{W}}}({\bf{I}_{{\rm{detail\_i}}}}) - {\bf{J}_{{\rm{detail\_i}}}})^T},
\end{align}
where $\alpha$ is the learning rate and $({\bf{I}_{{\rm{detail\_i}}}}, {\bf{J}_{{\rm{detail\_i}}}})$  is the $i$th patch pair.

\subsection{Combining CNN with image enhancement}
\label{sec.enhancement}
\begin{figure}[t!]
\centering
\subfigure[Rainy image]{
\includegraphics[width = 0.152\textwidth]{results/enhancement/1.jpg}}
\subfigure[Result by Eq. (\ref{eq.reconstruct})]{
\includegraphics[width = 0.152\textwidth]{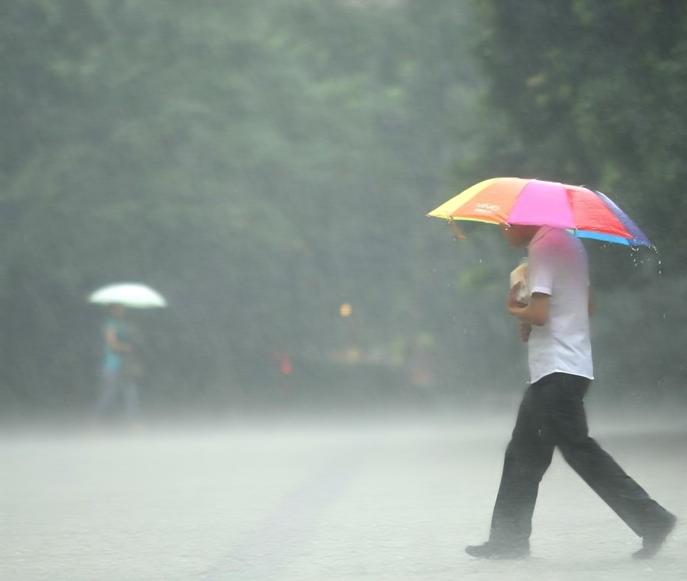}}
\subfigure[Result by Eq. (\ref{eq.recons_enahnced})]{
\includegraphics[width = 0.152\textwidth]{results/enhancement/1_proposedenhanced.jpg}}
\subfigure[Detail layer]{
\includegraphics[width = 0.2325\textwidth]{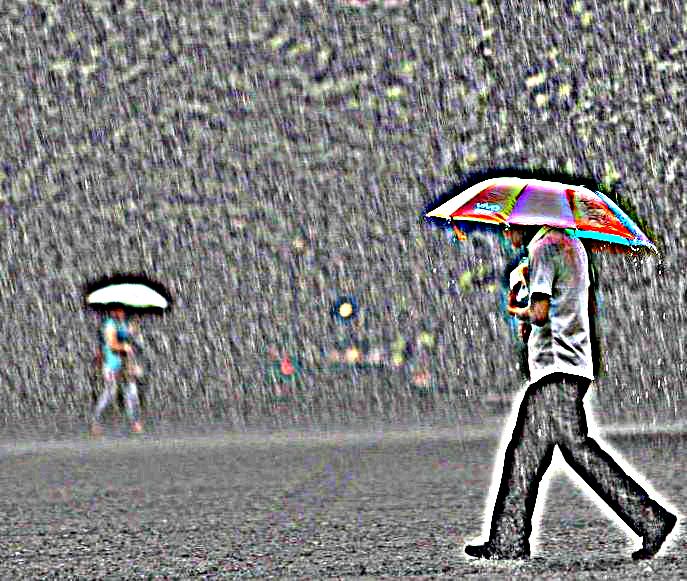}}
\subfigure[De-rained (d)]{
\includegraphics[width = 0.2325\textwidth]{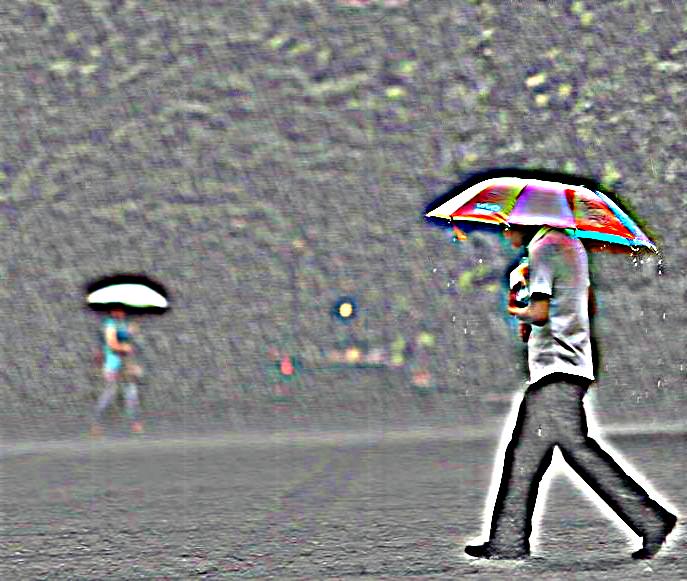}}
\subfigure[Base layer]{
\includegraphics[width = 0.2325\textwidth]{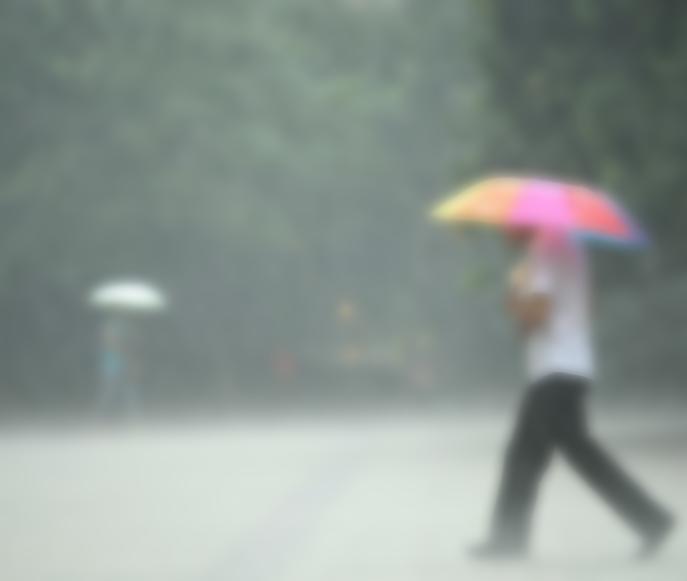}}
\subfigure[Enhanced (f)]{
\includegraphics[width = 0.2325\textwidth]{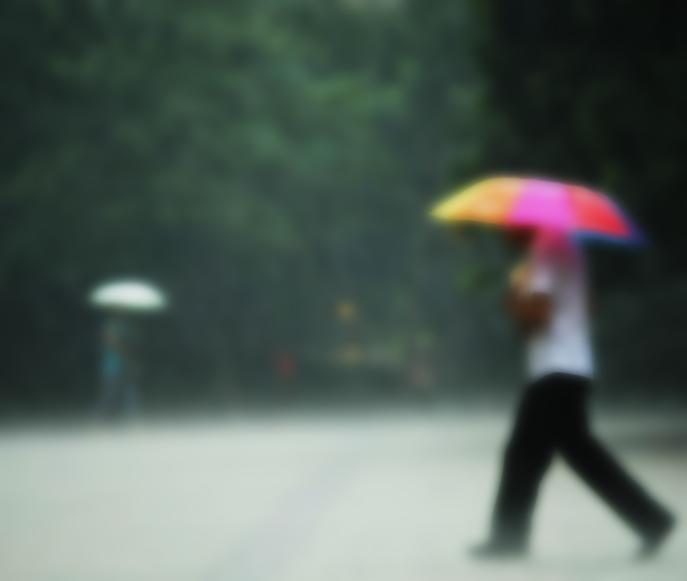}}
\caption{Visualization of intermediate results with enhancement. Intensities of detail layers have been amplified for better visualization.} \label{fig.enhance_intermediate}
\end{figure}

After training the network, the de-rained image can be obtained by directly adding the output detail layer to the base layer,
\begin{align}
\label{eq.reconstruct}
{\bf{O}} = {{\bf{I}}_{\rm{base}}}+{f_{\bf{w}}}({{\bf{I}}_{{\rm{detail}}}}),
\end{align}
where $\bf{O}$ is the de-rained output. However, when dealing with heavy rain the result unsurprisingly looks hazy, as shown in Figure \ref{fig.enhance_intermediate}(b). Fortunately, we can easily embed image enhancement technology into our framework to create a better visual result. Different mature and advanced image enhancement algorithms can be directly adopted in this framework as post-processing. In this paper, we use the non-linear function \cite{28} to enhance the base layer, and boost the detail layer by simply multiplying the output of the CNN by two to magnify the details,
\begin{align}
\label{eq.recons_enahnced}
{{\bf{O}}_{{\rm{enhanced}}}} = {({{\bf{I}}_{{\rm{base}}}})_{{\rm{enhanced}}}} + 2{f_{\bf{w}}}({{\bf{I}}_{{\rm{detail}}}}),
\end{align}
where ${{\bf{O}}_{{\rm{enhanced}}}}$ is the de-rained output with enhancement and ${({{\bf{I}}_{{\rm{base}}}})_{{\rm{enhanced}}}}$ is the enhanced base layer. Figure \ref{fig.enhance_intermediate}(c) shows the de-rained result with image enhancement. As shown in the intermediate results in Figures \ref{fig.enhance_intermediate}(d)-(g), virtually all of rain removal is being performed on the detail layer by the CNN, while the image enhancement on the base layer improves the global contrast and leads to a better visual result than shown in Figure \ref{fig.enhance_intermediate}(b) without using enhancement.


\section{Experiments}
To evaluate our DerainNet framework, we test on both synthetic and real-world rainy images. As mentioned previously, both testing frameworks are performed using the network trained on synthesized rainy images. We compare with three recent high quality de-raining methods \cite{13,16,34}. Software implementations of these methods were provided in Matlab by the authors. We use the default parameters reported in these three papers. All experiments are performed on a PC with Intel Core i5 CPU 4460, 8GB RAM and NVIDIA Geforce GTX 750. Our network contains two hidden layers and one output layer as described in Section \ref{sec.ourCNN}. We set kernel sizes $s_1 = 16, s_2 = 1$ and $s_3 = 8$, respectively. The number of feature maps for each hidden layer are $n_1 = n_2 = 512$. We set the learning rate to $\alpha  = 0.01$. More visual results and our Matlab implementation can be found at \url{http://smartdsp.xmu.edu.cn/derainNet.html}.
\begin{figure}
\subfigure[Ground truth]{
\begin{minipage}[t]{0.313\columnwidth}
\centering
\includegraphics[width=1\textwidth]{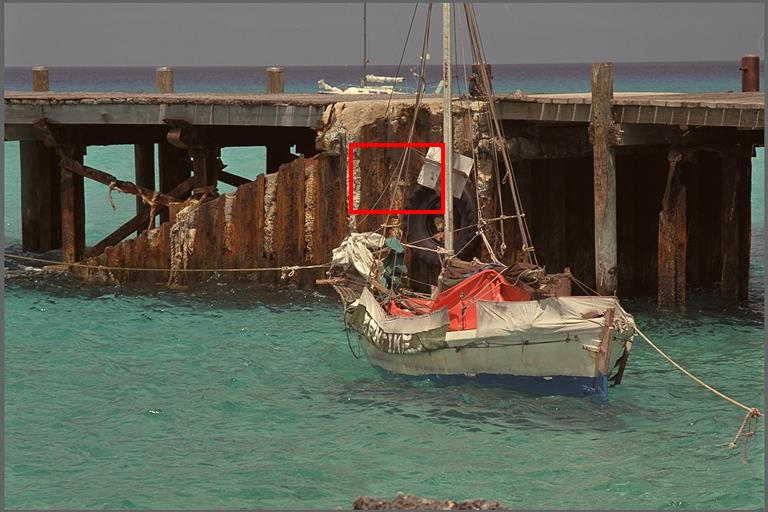}\\ \vspace{1.5pt}
\includegraphics[width=.7\textwidth]{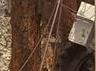}\vspace{1pt}
\end{minipage}}
\subfigure[Rainy image]{
\begin{minipage}[t]{0.313\columnwidth}
\centering
\includegraphics[width=1\textwidth]{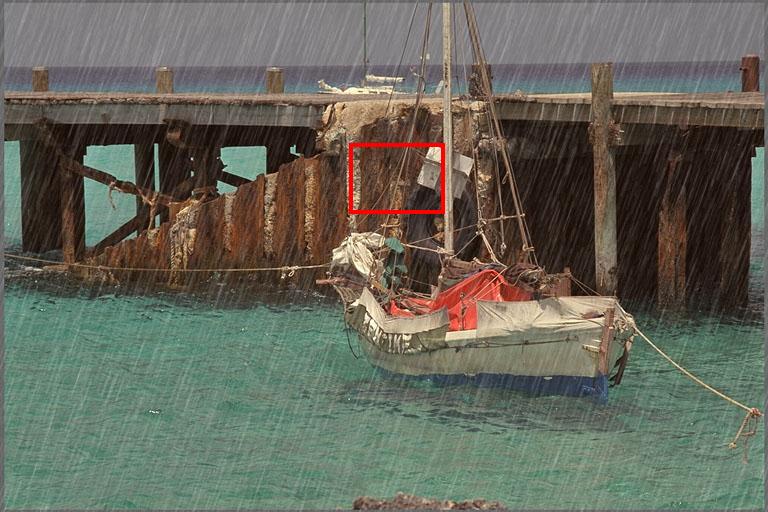} \\ \vspace{1.5pt}
\includegraphics[width=.7\textwidth]{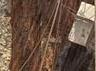}\vspace{1pt}
\end{minipage}}
\subfigure[Method \cite{13}]{
\begin{minipage}[t]{0.313\columnwidth}
\centering
\includegraphics[width=1\textwidth]{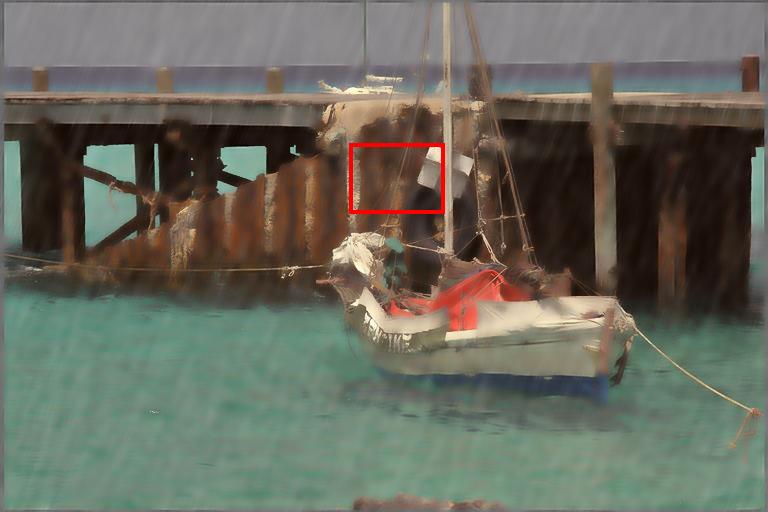} \\ \vspace{1.5pt}
\includegraphics[width=.7\textwidth]{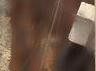}\vspace{1pt}
\end{minipage}}
\subfigure[Method \cite{16}]{
\begin{minipage}[t]{0.313\columnwidth}
\centering
\includegraphics[width=1\textwidth]{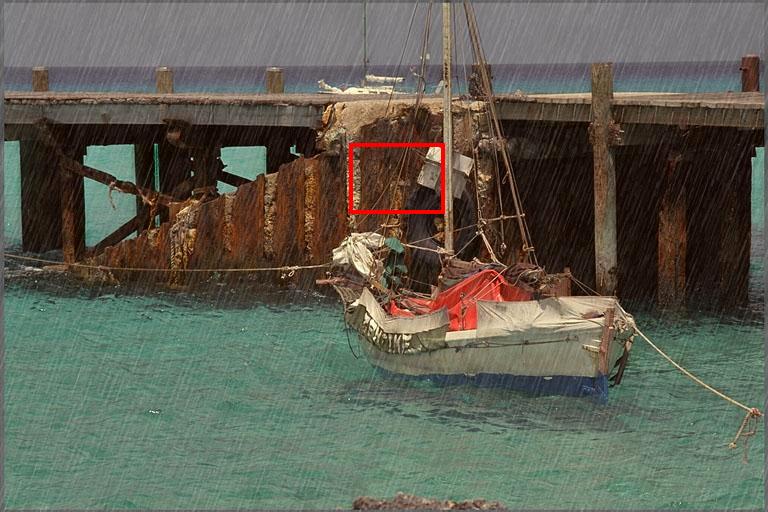} \\ \vspace{1.5pt}
\includegraphics[width=.7\textwidth]{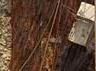}\vspace{1pt}
\end{minipage}}
\subfigure[Method \cite{34}]{
\begin{minipage}[t]{0.313\columnwidth}
\centering
\includegraphics[width=1\textwidth]{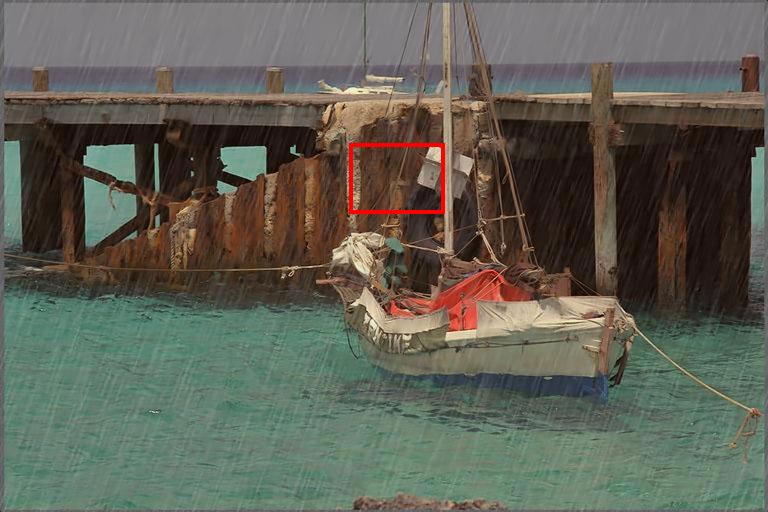} \\ \vspace{1.5pt}
\includegraphics[width=.7\textwidth]{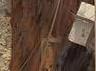}\vspace{1pt}
\end{minipage}}
\subfigure[Our result]{
\begin{minipage}[t]{0.313\columnwidth}
\centering
\includegraphics[width=1\textwidth]{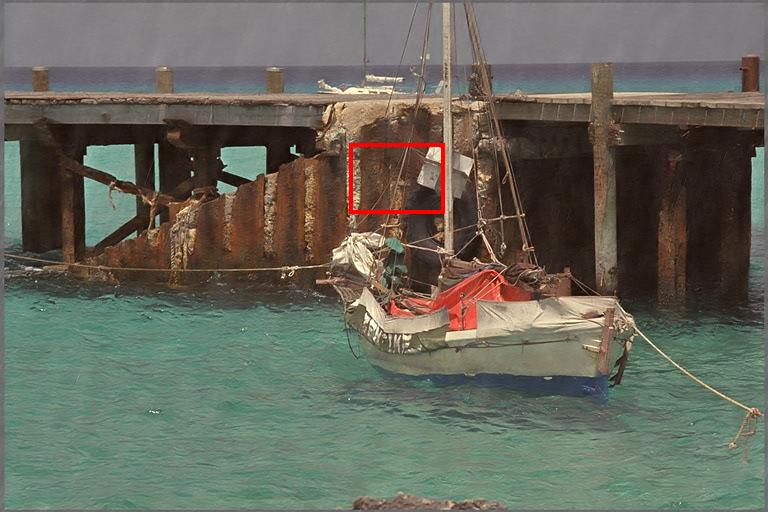} \\ \vspace{1.5pt}
\includegraphics[width=.7\textwidth]{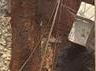}\vspace{1pt}
\end{minipage}}
\caption{Results on synthesized rainy image ``dock''. Row 2 shows corresponding enlarged parts of red boxes in Row 1.} \label{fig.synthetic_enlarge}
\end{figure}

\begin{figure*}
\centering
\includegraphics[width=.161\textwidth]{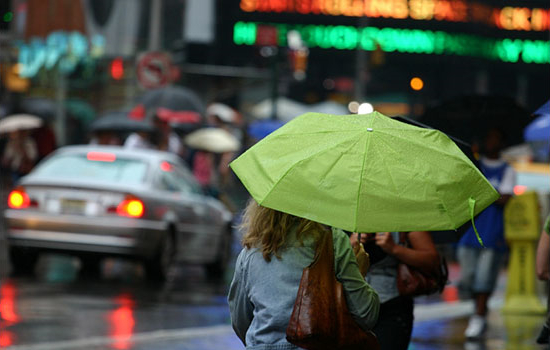}\vspace{1pt}
\includegraphics[width=.161\textwidth]{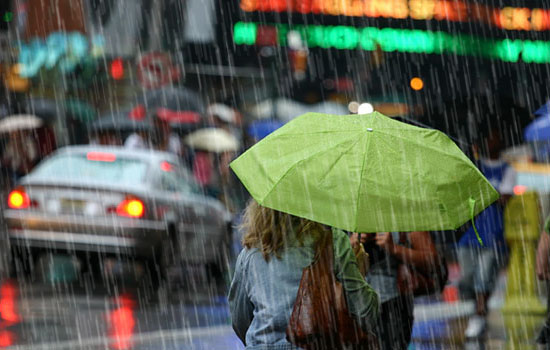}\vspace{1pt}
\includegraphics[width=.161\textwidth]{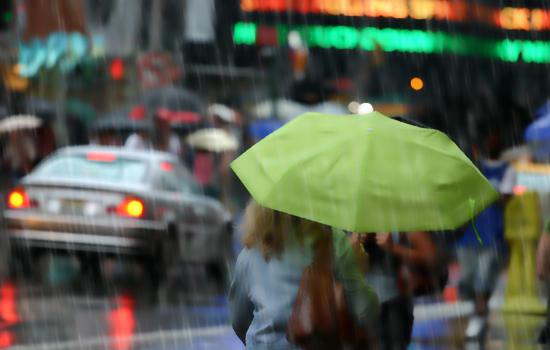}\vspace{1pt}
\includegraphics[width=.161\textwidth]{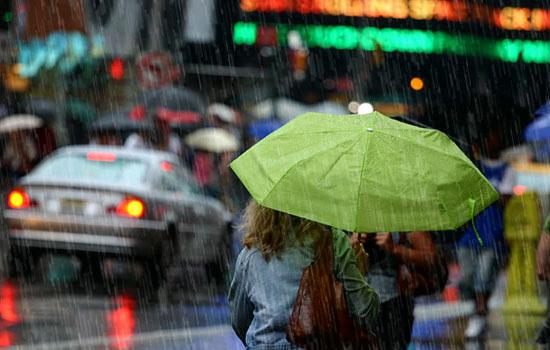}\vspace{1pt}
\includegraphics[width=.161\textwidth]{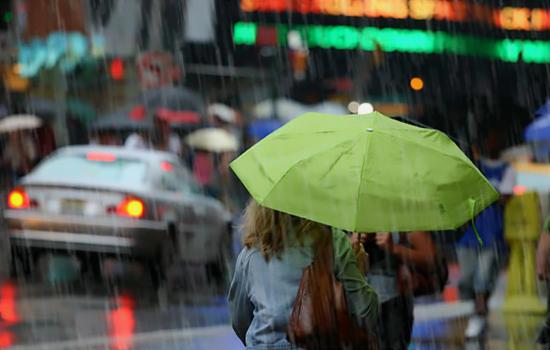}\vspace{1pt}
\includegraphics[width=.161\textwidth]{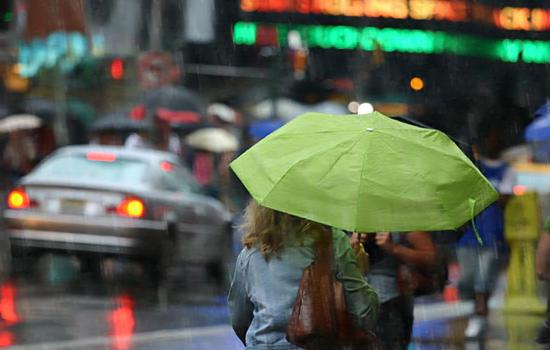}\vspace{1pt}

\includegraphics[width=.161\textwidth]{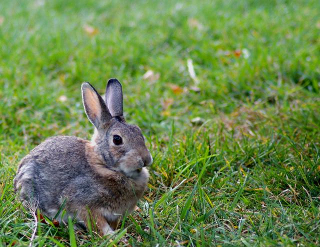}\vspace{1pt}
\includegraphics[width=.161\textwidth]{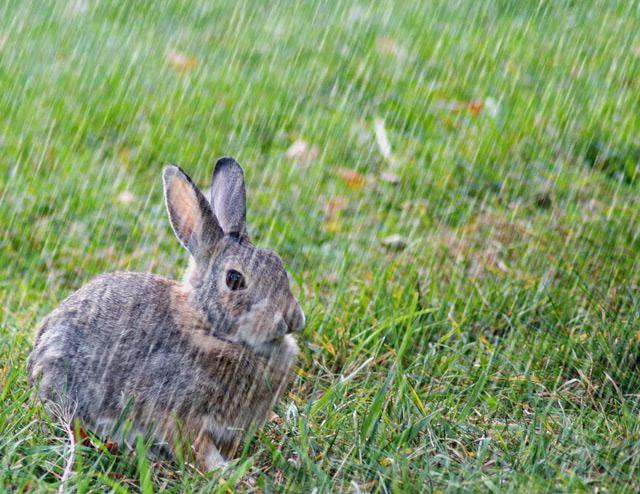}\vspace{1pt}
\includegraphics[width=.161\textwidth]{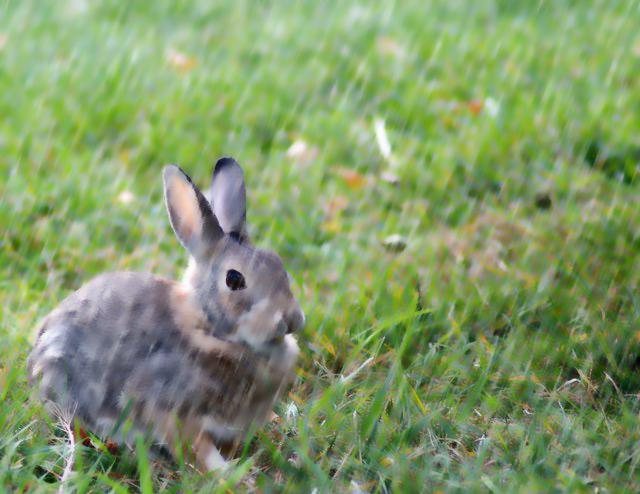}\vspace{1pt}
\includegraphics[width=.161\textwidth]{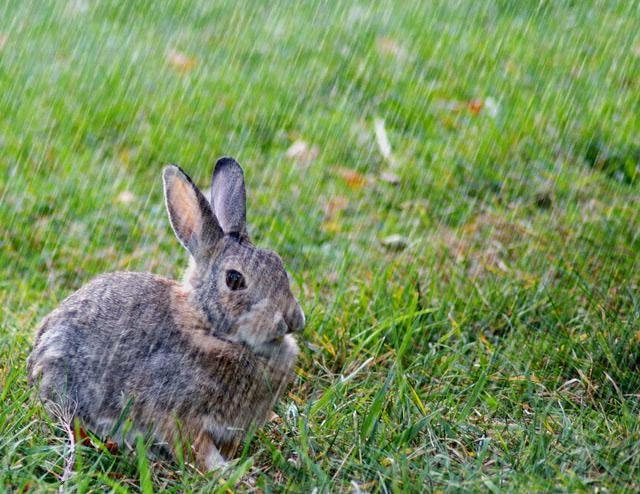}\vspace{1pt}
\includegraphics[width=.161\textwidth]{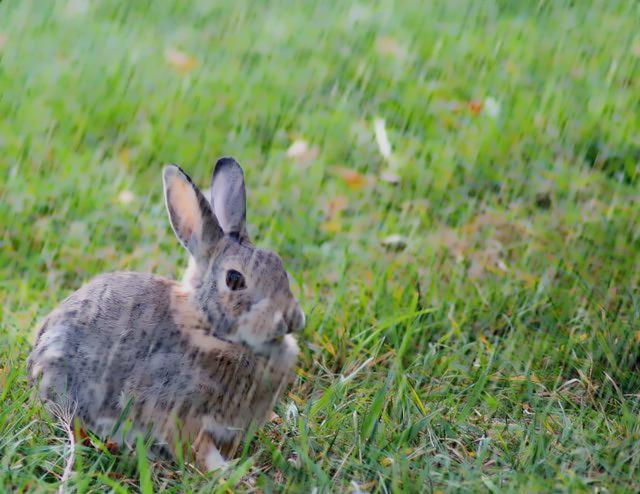}\vspace{1pt}
\includegraphics[width=.161\textwidth]{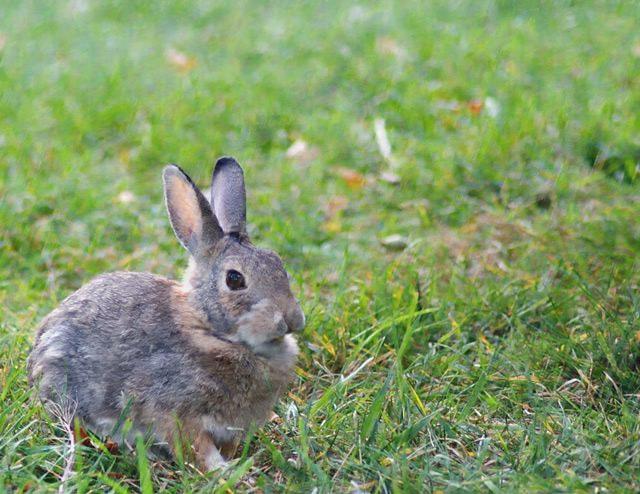}\vspace{1pt}

\includegraphics[width=.161\textwidth]{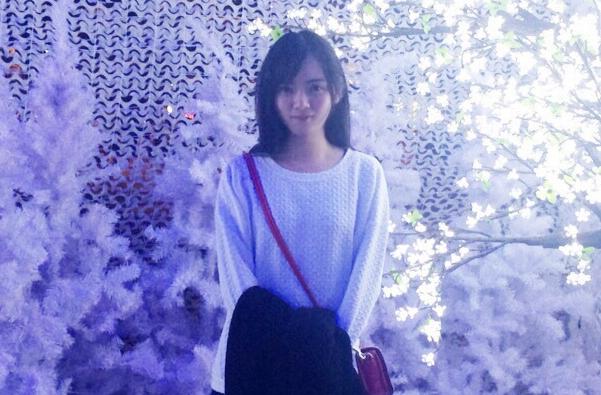}
\includegraphics[width=.161\textwidth]{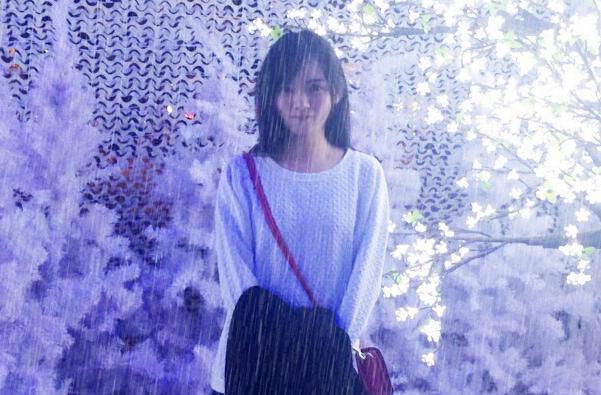}
\includegraphics[width=.161\textwidth]{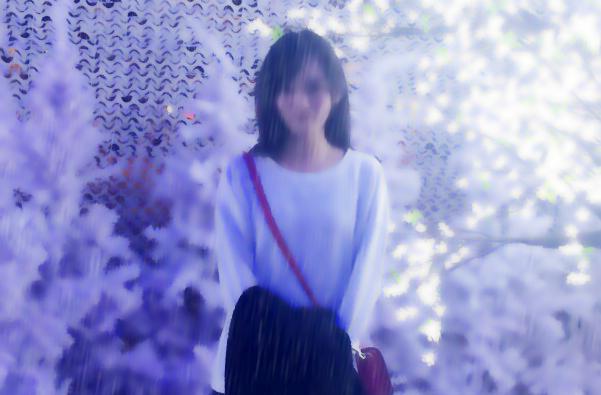}
\includegraphics[width=.161\textwidth]{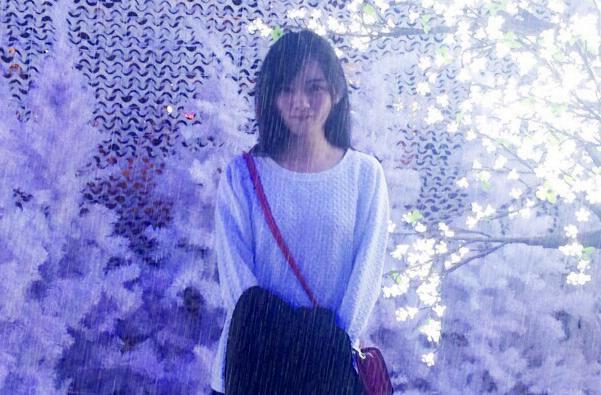}
\includegraphics[width=.161\textwidth]{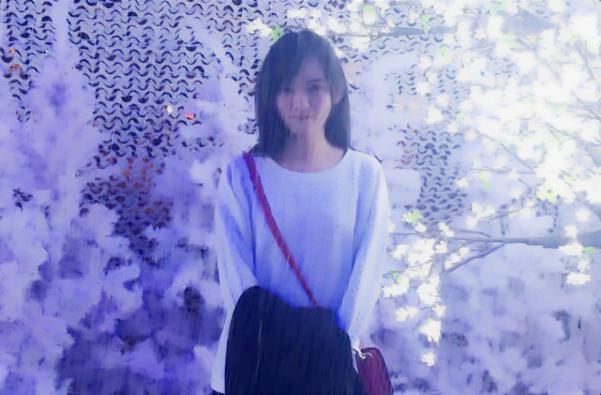}
\includegraphics[width=.161\textwidth]{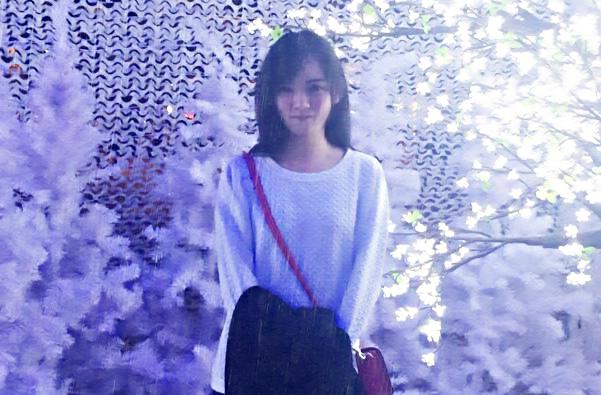}

\subfigure[Ground truth]{\includegraphics[width=.161\textwidth]{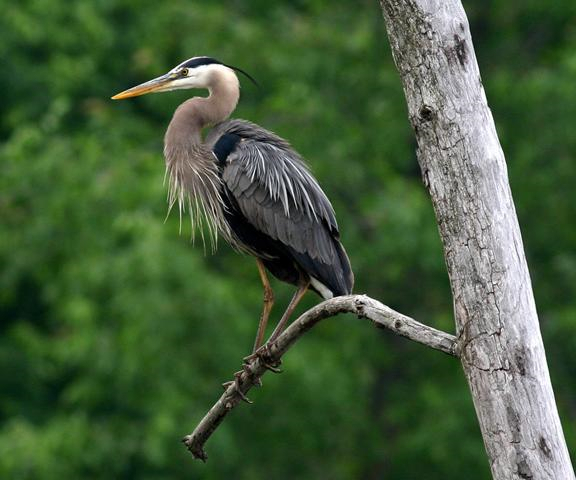}}
\subfigure[Synthesized image]{\includegraphics[width=.161\textwidth]{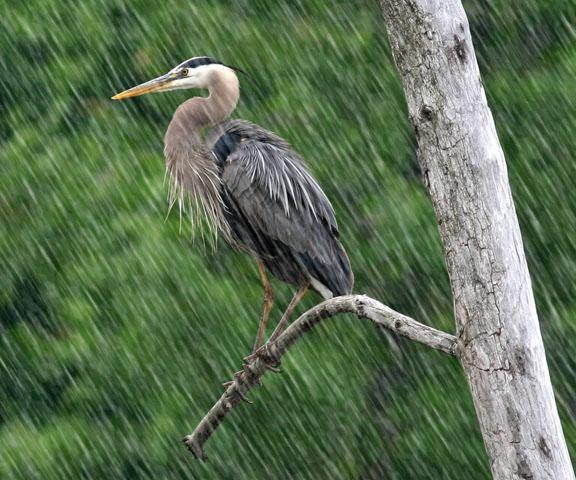}}
\subfigure[Method \cite{13}]{\includegraphics[width=.161\textwidth]{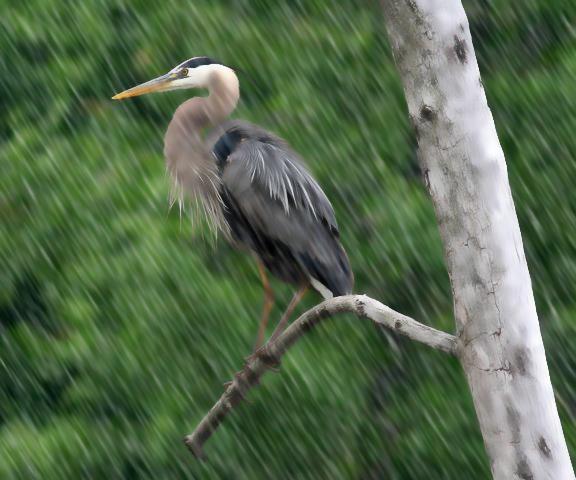}}
\subfigure[Method \cite{16}]{\includegraphics[width=.161\textwidth]{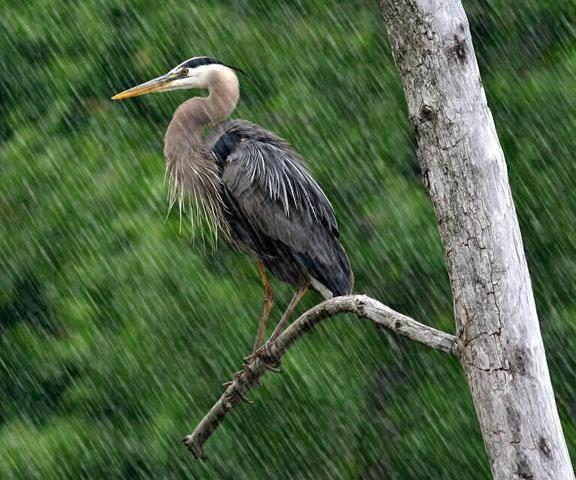}}
\subfigure[Method \cite{34}]{\includegraphics[width=.161\textwidth]{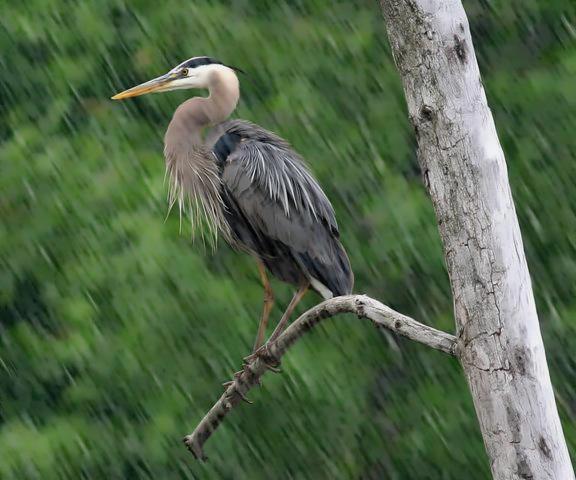}}
\subfigure[DerainNet]{\includegraphics[width=.161\textwidth]{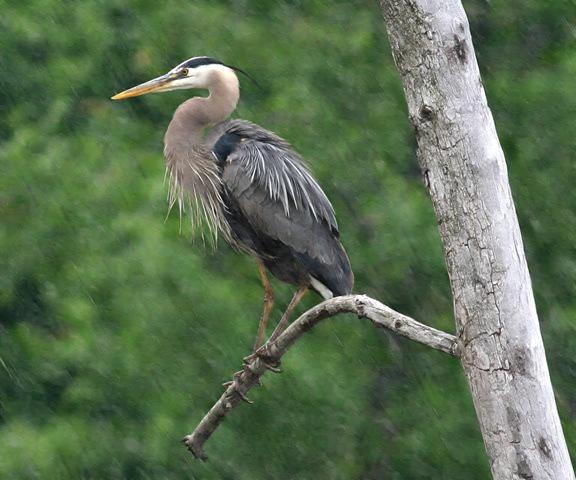}}
\caption{Example results on synthesized rainy images ``umbrella'', ``rabbit'', ``girl'' and ``bird.'' These rainy images were for testing and not used for training.} \label{fig.synthetic1}
\end{figure*}

\subsection{Synthesized data}
We first evaluate the results of testing on newly synthesized rainy images. In our first results, we synthesize new rainy images by selecting from the set of 350 clean images from our database. Figure \ref{fig.synthetic_enlarge}  shows visual comparisons for one such synthesized test image. As can be seen, method \cite{13} exhibits over-smoothing of the rope and method \cite{16,34} leaves significant rain streaks in the result. This is because \cite{13,16,34} are algorithms based on low-level image features. When the rope's orientation and magnitude is similar with that of rain, methods \cite{13,16,34} cannot efficiently distinguish the rope from rain streaks. However, as shown in the last result, the multiple convolutional layers of DerainNet can identify and remove rain while preserving the rope.

Figure \ref{fig.synthetic1}  shows visual comparisons for four more synthesized rainy image using different rain streak orientations and magnitudes. Since the ground truth is known, we use the the structure similarity index (SSIM) \cite{30} for quantitative evaluation. A higher SSIM value indicates a de-rained image that is closer to the ground truth in terms of image structural properties. (For the ground truth, the SSIM equals 1.) For a fair comparison, the image enhancement operation is \textit{not} implemented by our algorithm for these synthetic experiments.
\begin{table*}
\caption{Quantitative measurement results using SSIM on synthesized test images.}
\label{tab.SSIM}
\centering
\begin{tabular}{|c|c|c|c|c|c|c|}
\hline
Images &Ground truth &Rainy image& Method \cite{13} & Method \cite{16} & Method \cite{34} &Ours\\
\hline
dock&1&  0.86 & 0.84 &   0.88  & 0.90 & \textbf{0.92}\\
\hline
umbrella&1& 0.75 &0.83 &   0.81 &0.86 &\textbf{0.88}\\
\hline
rabbit&1& 0.72 &0.74  &  0.77  &  0.79&\textbf{0.85}\\
\hline
girl&1& 0.93&0.80  &  \textbf{0.94}  & 0.91 &\textbf{0.94}\\
\hline
bird&1& 0.57&0.63  &  0.62  & 0.75 &\textbf{0.82}\\
\hline
100 new images&1 & 0.79 $\pm$ 0.13  & 0.73 $\pm$ 0.07  &  0.84 $\pm$ 0.09 & 0.82 $\pm$ 0.10  & \textbf{0.89 $\pm$ 0.06}\\
\hline
\emph{Rain12} \cite{34}&1 & 0.91 $\pm$ 0.05  & 0.81 $\pm$ 0.07  &  0.88 $\pm$ 0.05 & 0.91 $\pm$ 0.03 & \textbf{0.92 $\pm$ 0.03}\\
\hline
\end{tabular}
\end{table*}

As is again evident in these results, method \cite{13} over-smooths the results and methods \cite{16,34} leave rain streaks, both of which are addressed by our algorithm. Moreover, we see in Table \ref{tab.SSIM} that our method has the highest SSIM values, in agreement with the visual effect. Also shown in Table \ref{tab.SSIM} is the performance of the three methods on 100 newly-synthesized testing images using our synthesizing strategy. In Table \ref{tab:matrix} we show the number of images for which the algorithm on the row outperformed the algorithm on the column for these 100 images.

In Table \ref{tab.SSIM} we also show results applying the same trained algorithms for each method on $12$ newly synthesized rainy images (called \emph{Rain12}) \cite{34} that are generated using photorealistic rendering techniques \cite{2}. This clearly highlights the generalizability of DerainNet to new scenes; whereas the other algorithms either decrease the performance or leave it unchanged, DerainNet still shows improvement.

\begin{table}
\centering
\caption{$\#$ times (row) beat (col)}\label{tab:matrix}
\begin{tabular}{|r|c|c|c|c|} \hline
 & \cite{13} & \cite{16} & \cite{34} &Ours \\ \hline
\cite{13} & $-$ & 5 & 4& 0\\   \hline
\cite{16} &107 & $-$ &76 & 6\\   \hline
\cite{34} & 108 & 36&$-$ &7 \\  \hline
Ours & 112 & 102 & 105&$-$\\   \hline
\end{tabular}
\end{table}

\subsection{Real-world data}
\begin{figure*}
\begin{center}
\subfigure[Rainy image]{\includegraphics[width = .18\textwidth]{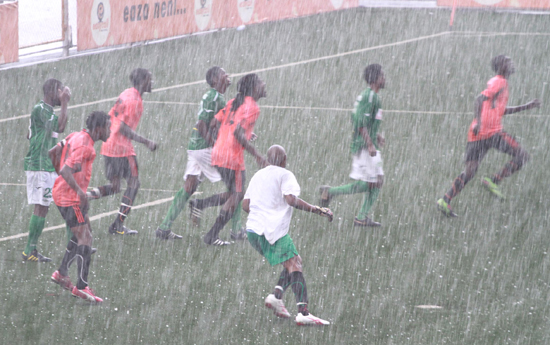}}
\subfigure[Method \cite{13}]{\includegraphics[width = .18\textwidth]{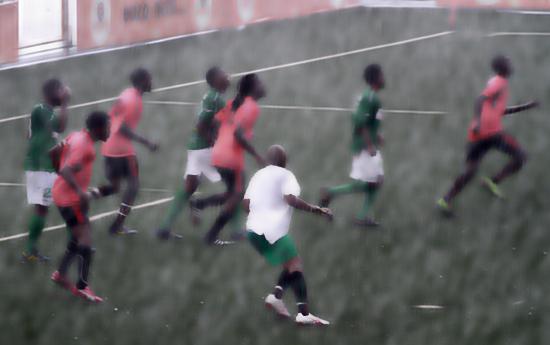}}
\subfigure[Method \cite{16}]{\includegraphics[width = .18\textwidth]{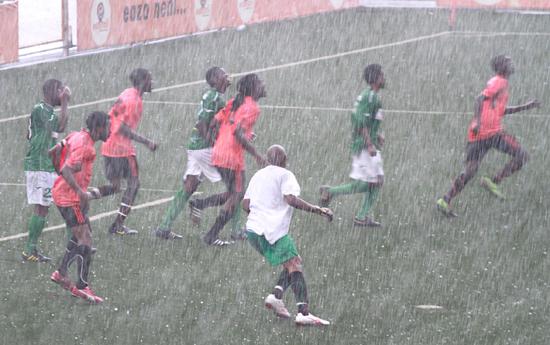}}
\subfigure[Method \cite{34}]{\includegraphics[width = .18\textwidth]{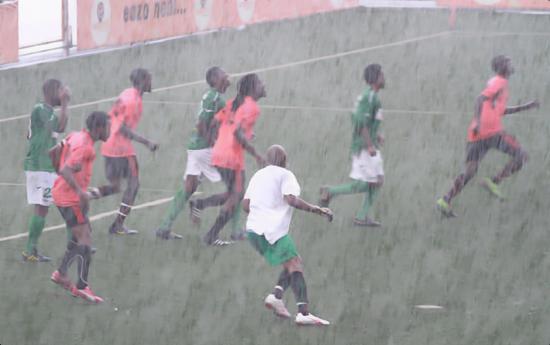}}
\subfigure[Our result]{\includegraphics[width = .18\textwidth]{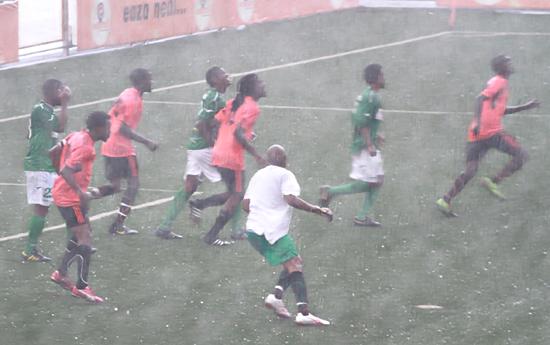}}\\
\subfigure[Rainy enhanced]{\includegraphics[width = .18\textwidth]{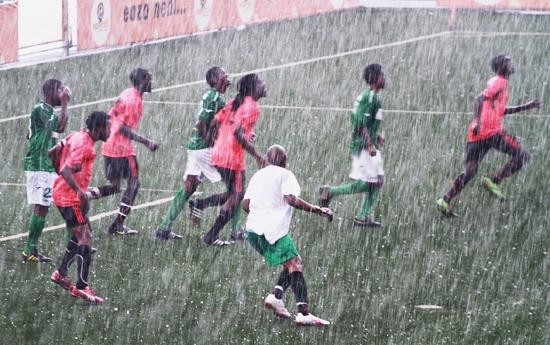}}
\subfigure[Method \cite{13} enhanced]{\includegraphics[width = .18\textwidth]{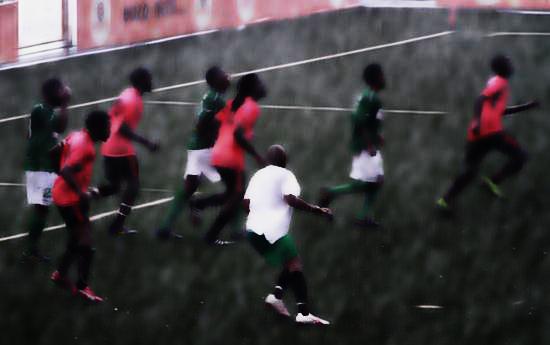}}
\subfigure[Method \cite{16} enhanced]{\includegraphics[width = .18\textwidth]{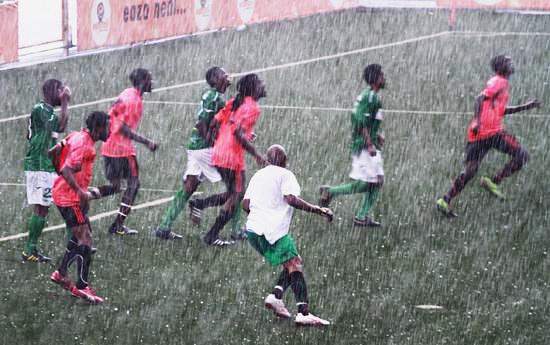}}
\subfigure[Method \cite{34} enhanced]{\includegraphics[width = .18\textwidth]{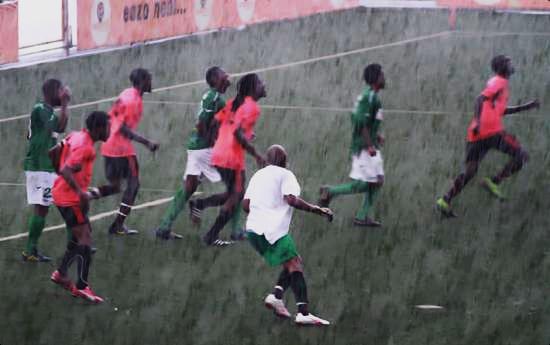}}
\subfigure[Our result enhanced]{\includegraphics[width = .18\textwidth]{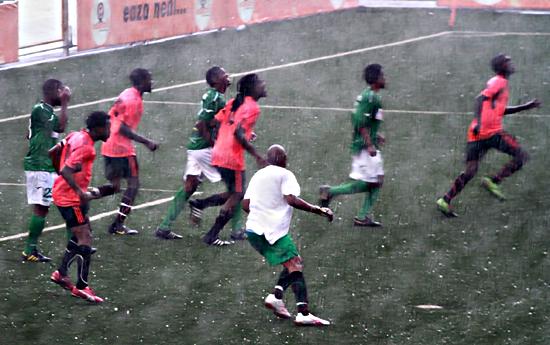}}
\end{center}\vspace{-5pt}
\caption{Comparison of algorithms on a real-world ``soccer'' image with and without enhancement.}\label{fig.enhanced}
\end{figure*}

\begin{figure*}
\centering
\includegraphics[width = .18\textwidth]{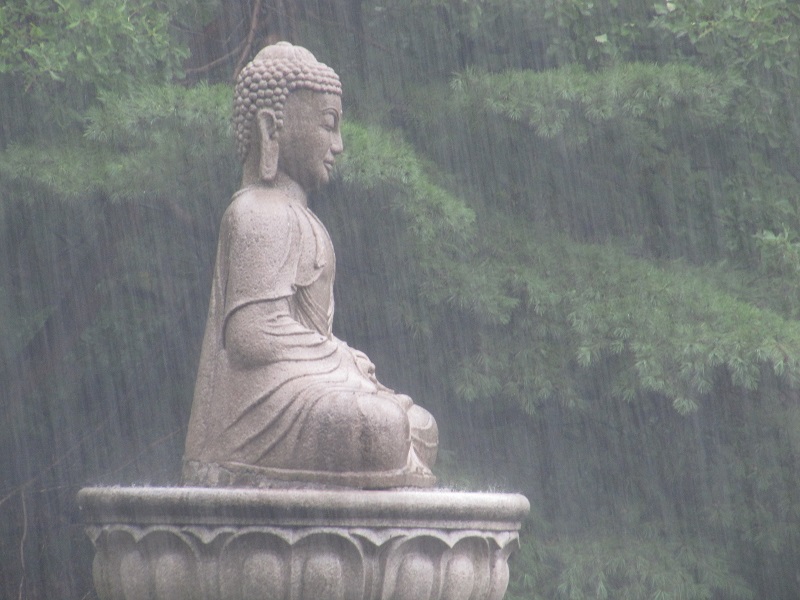}
\includegraphics[width = .18\textwidth]{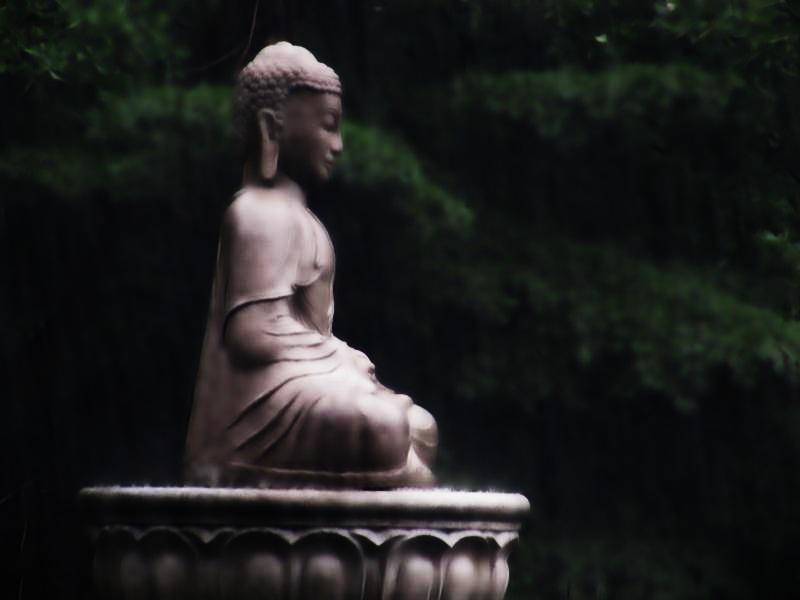}
\includegraphics[width = .18\textwidth]{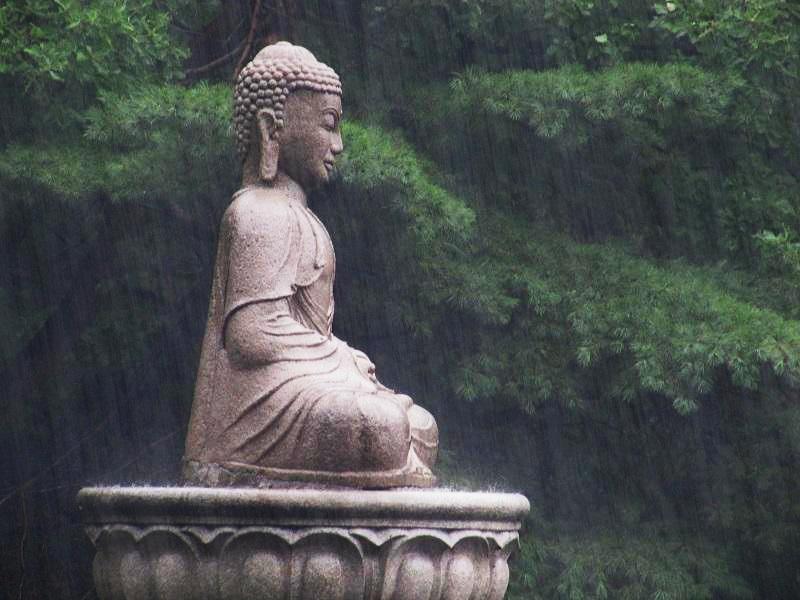}
\includegraphics[width = .18\textwidth]{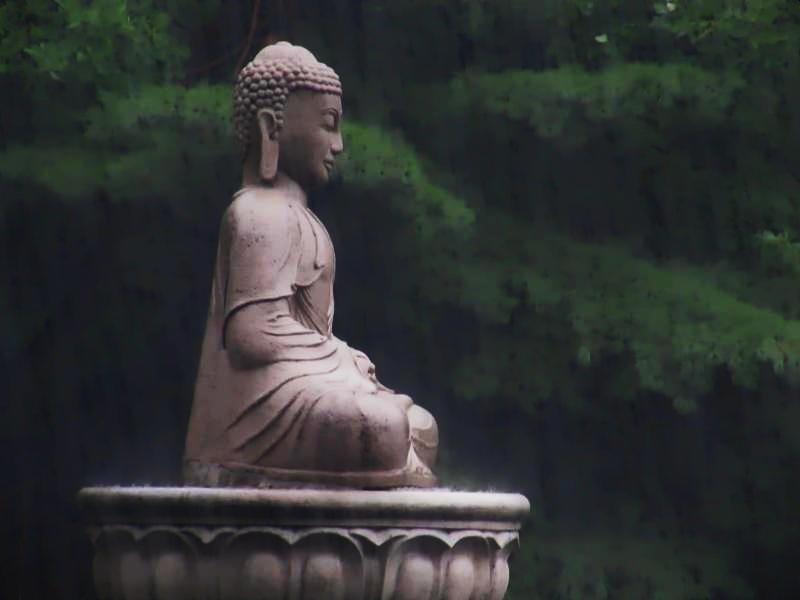}
\includegraphics[width = .18\textwidth]{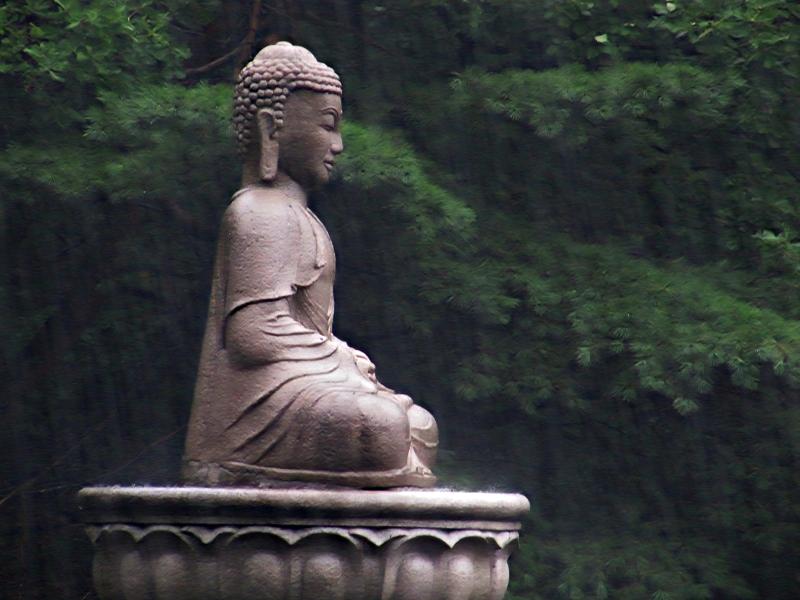}\vspace{4pt} \\

\includegraphics[width = .18\textwidth]{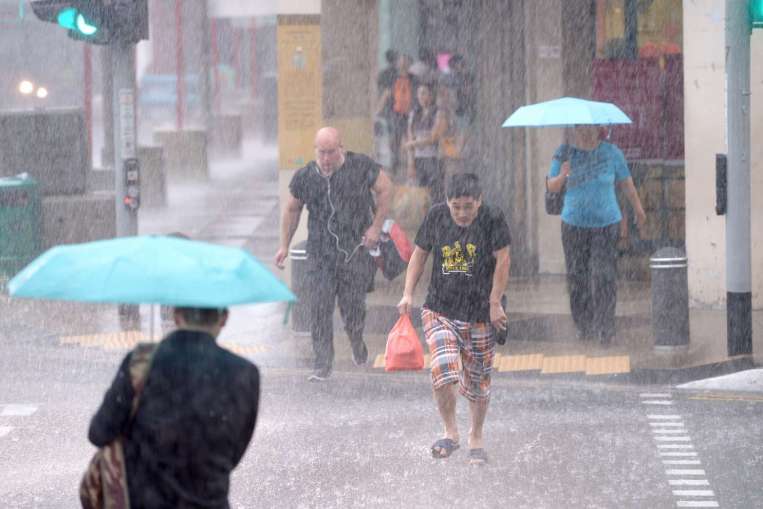}
\includegraphics[width = .18\textwidth]{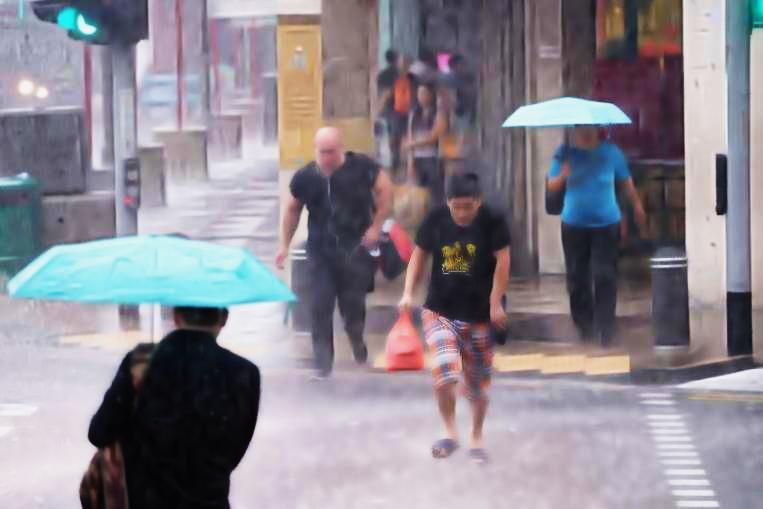}
\includegraphics[width = .18\textwidth]{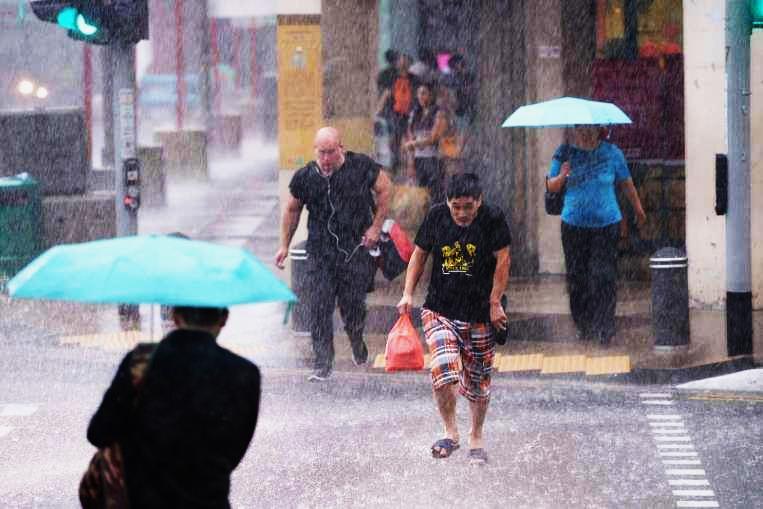}
\includegraphics[width = .18\textwidth]{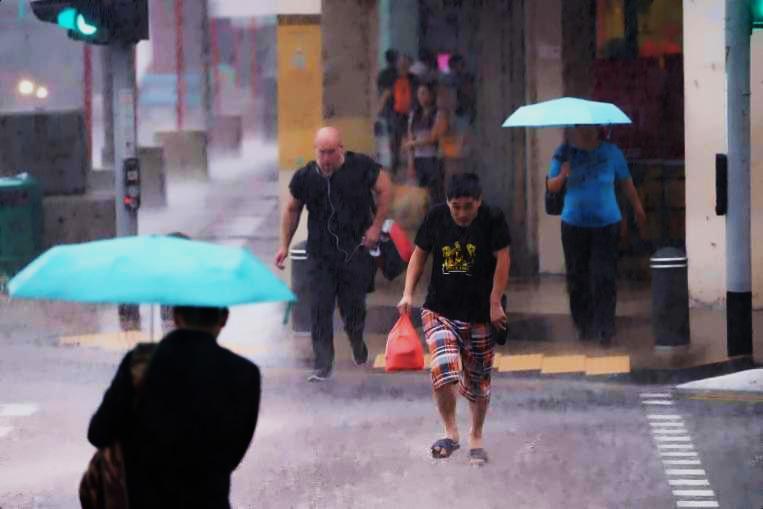}
\includegraphics[width = .18\textwidth]{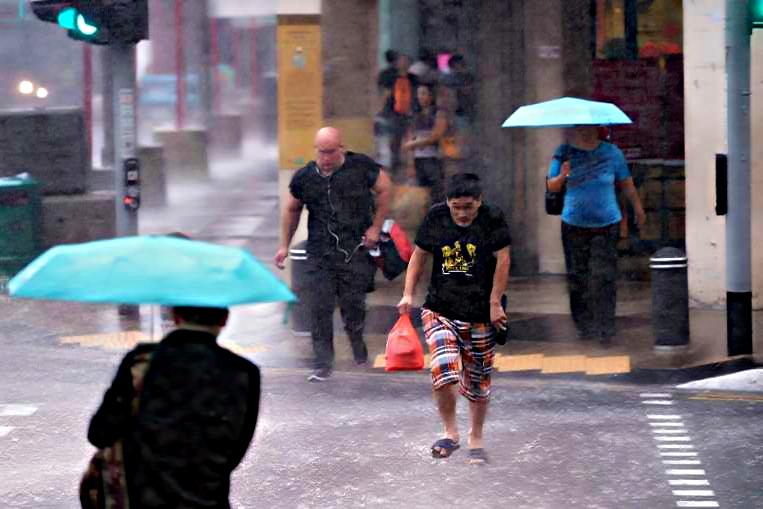}\vspace{-1pt} \\

\subfigure[Rainy images]{\includegraphics[width = .18\textwidth]{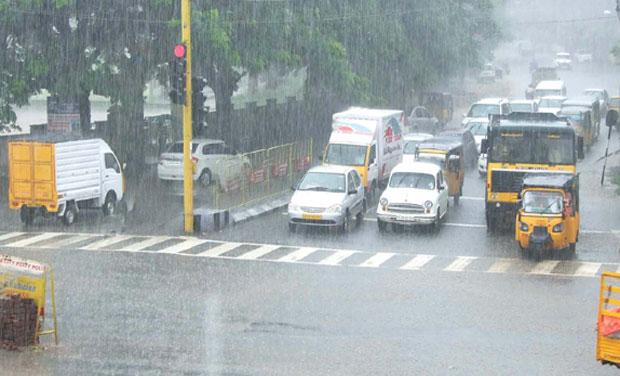}}
\subfigure[Method \cite{13}]{\includegraphics[width = .18\textwidth]{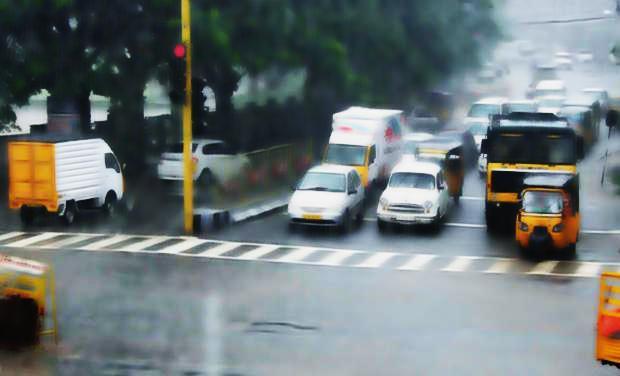}}
\subfigure[Method \cite{16}]{\includegraphics[width = .18\textwidth]{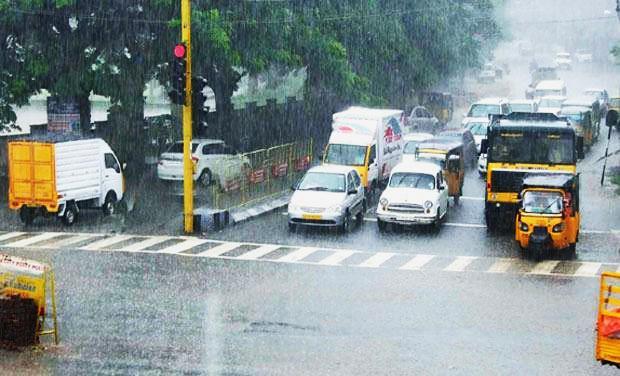}}
\subfigure[Method \cite{34}]{\includegraphics[width = .18\textwidth]{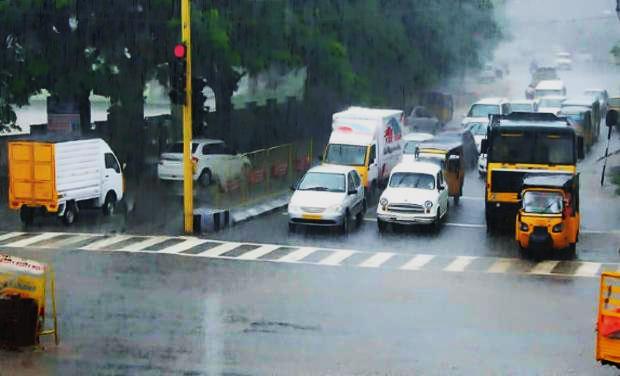}}
\subfigure[Our results]{\includegraphics[width = .18\textwidth]{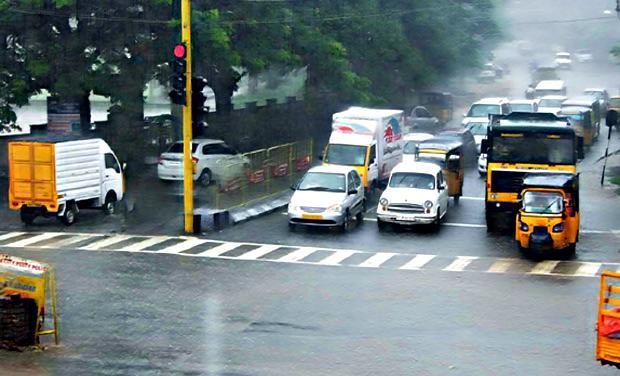}}
\caption{Three more results on real-world rainy images: (top-to-bottom) ``Buddha,'' ``street,'' ``cars.'' All algorithms use image enhancement.} \label{fig.real3}
\end{figure*}

Since we do not possess the ground truth corresponding to real-world rainy images, we test DerainNet on real-world data using the network trained on the $4900$ synthesized images from the previous section. In Figure \ref{fig.enhanced} we show the results of all algorithms with and without enhancement, where enhancement of \cite{13,16} and \cite{34} are performed as post-processing, and for DerainNet is performed as shown in Figure \ref{fig.overview}. In our quantitative comparison below, we use enhancement for all results, but note that the relative performance between algorithms was similar without using enhancement. We show results on three more real-world rainy images in Figure \ref{fig.real3}. Although we use synthetic data to train our DerainNet, we see that this is sufficient for learning a network that is effective when applied to real-world images.

\begin{table*}[ht]
\caption{Quantitative measurement results of BIQI on real-world test images.}
\label{tab.NIQE}
\centering
\begin{tabular}{|c|c|c|c|c|c|}
\hline
Images &Input & Method \cite{13} & Method \cite{16} & Method \cite{34} &Ours\\
\hline
soccer&57.96& 42.70 &   53.86 &35.64 &\textbf{33.35}\\
\hline
Buddha&49.06& 39.55 &  50.13  & 39.90 &\textbf{28.10}\\
\hline
street&37.70& 40.67  &  38.32 & 38.98 &\textbf{34.08}\\
\hline
cars & 27.84 & 40.08 & \textbf{21.17} &31.70 &24.18\\
\hline
100 test images&33.00 $\pm$ 13.19& 39.63 $\pm$ 7.66  & 31.43 $\pm$ 9.81 &  34.60 $\pm$ 7.78 &\textbf{29.86 $\pm$ 6.98}\\
\hline
\end{tabular}
\end{table*}
In Figure \ref{fig.real3}, the proposed method arguably shows the best visual performance on simultaneously removing rain and preserving details. Since the ground truth is unavailable in these examples, we cannot definitively say which algorithm performs quantitatively the best. Instead, we use a reference-free measure called the Blind Image Quality Index (BIQI) \cite{moorthy2010two} for quantitative evaluation. This index is designed to provide a score of the quality of an image without reference to ground truth. A lower value of BIQI indicates a higher quality image. However, as with all reference-free image quality metrics, BIQI is arguably not always subjectively correct. Still, as Table \ref{tab.NIQE} indicates, our method has the lowest BIQI on 100 newly obtained real-world testing images. This gives additional evidence that our method outputs an image with greater improvement.


To provide realistic feedback and quantify the subjective evaluation of DerainNet, we also constructed an independent user study. In this experiment, we use the de-rained results (with enhancement) of the same 100 real-world images scored with BIQI. For each image, we randomly order the outputs of the four algorithms, as well as the original rainy image, and display them on a screen. We then separately asked 20 participants to rank each image from 1 to 5 subjectively according to quality, with the instructions being that visible rain should decrease the quality and clarity should increase quality (1 represents the worst quality image and 5 represents the best quality image).  We show the average scores in Table \ref{tab.User} from these 2000 trials. As is evident, methods \cite{13} and \cite{16} do not make a clear improvement over the original image. Method \cite{36} does clearly improve the rainy image, but the proposed method is subjectively superior to all images. This small-scale experiment gives 
additional support along with BIQI and our own subjective assessment that DerainNet improves the de-raining on real-world images.
\begin{table}
\caption{Average scores of user study.}
\label{tab.User}
\centering
\begin{tabular}{|c|c|c|c|c|c|}
\hline
Images &Input & Method \cite{13} & Method \cite{16} & Method \cite{34} &Ours\\
\hline
Scores&1.51 & 1.46  & 1.73 & 2.57 &\textbf{4.11}\\
\hline
\end{tabular}
\end{table}


\subsection{Parameter settings}
In this section, we test different parameters setting to study their impact on performance. We use the same training data as previously. The testing data includes the same 100 newly-synthesized images as well as the new \emph{Rain12} images \cite{34}.

\subsubsection{Kernel size}
First, we test the impact of different kernel sizes. The default kernel sizes for the three levels are 16, 1 and 8; we denote this network as 16-1-8. We fix the kernel size of the second layer and reduce the kernel sizes of first and third layers to 4-1-2 and 8-1-4. We then performed experiments by instead increasing the kernel size of second layer to 16-3-8 and 16-5-8. Table \ref{tab.Kernel_size} shows the average SSIM values for these different kernel sizes. As can be seen, larger kernel sizes can generate better results. This is because more structure and texture can be modeled using a large kernel. On the contrary, from our experiments we find that increasing the kernel size of the second layer brings only limited improvement. This is because the second layer performs a non-linear operation for rain removal and the $1 \times 1$ kernel can achieve promising results. Thus, we choose 16-1-8 as the default setting of kernel size.

\begin{table}[!h]
\caption{Average SSIM of different kernel sizes.}
\label{tab.Kernel_size}
\centering
\begin{tabular}{|c|c|c|c|c|c|}
\hline
Kernel sizes &4-1-2 &8-1-4 & 16-1-8 (default) \\
\hline
SSIM&0.84 $\pm$ 0.06 &  0.87 $\pm$ 0.05  &0.89 $\pm$ 0.06 \\
\hline
\end{tabular}
\begin{tabular}{|c|c|c|c|c|c|}
\hline
Kernel sizes &16-3-8 & 16-5-8   \\
\hline
SSIM&0.89 $\pm$ 0.06 &  0.90 $\pm$ 0.07    \\
\hline
\end{tabular}
\end{table}

\subsubsection{Network width}
Intuitively, if we increase the network width by increasing the number of kernels, $n_1$ and $n_2$, the performance should improve. We train three models by using the values: $n_1, n_2 \in \{64, 128, 256\}$ and compare them to our default setting of $n_1 = n_2= 512$. Table \ref{tab.Width}  shows the average SSIM values for these four models. As can be seen, better performance can be achieved by increasing the width of the network. However, increasing the number of kernels improves the performance at the cost of running time since more convolutional operations are required. Thus we choose $n_1 = n_2= 512$ as the default setting of network width.

\begin{table}[!h]
\caption{Average SSIM of different network width.}
\label{tab.Width}
\centering
\begin{tabular}{|c|c|c|c|c|c|}
\hline
Width &64 & 128 & 256 & 512 (default) \\
\hline
SSIM&0.80 $\pm$ 0.05 &  0.82 $\pm$ 0.05 & 0.85 $\pm$ 0.05 & \textbf{0.89 $\pm$ 0.06} \\
\hline
\end{tabular}
\end{table}

\subsubsection{Network depth}
We also test the performance of using deeper structures by adding more non-linear layers. We train and test on 3 networks with depths 3, 5 and 10. As shown in Table \ref{tab.network_depth}, for the de-raining problem, increasing the network depth does not bring better results using a feed-forward network structure. This is a results of gradient vanishing, which may perhaps be addressed by designing a more complex network structure (with increased computation time). However, our DerainNet generates high quality results with only 3 layers as a result of our proposed detail training strategy, and so the complexity and computation time of the model can be significantly reduced. Therefore, we adopt three layers as the default setting.
\begin{table}[t]
\caption{Average SSIM of different network depth.}
\label{tab.network_depth}
\centering
\begin{tabular}{|c|c|c|c|c|c|}
\hline
Depth &3 (default)&5 & 10 \\
\hline
SSIM&\textbf{0.89 $\pm$ 0.06} &  0.84 $\pm$ 0.05  &0.79 $\pm$ 0.04 \\
\hline
\end{tabular}
\end{table}

\begin{figure}
{\includegraphics[width=.2425\columnwidth]{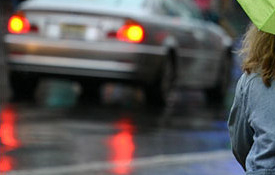}}
{\includegraphics[width=.2425\columnwidth]{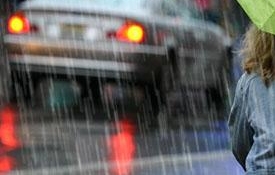}}
{\includegraphics[width=.2425\columnwidth]{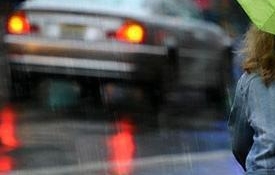}}
{\includegraphics[width=.2425\columnwidth]{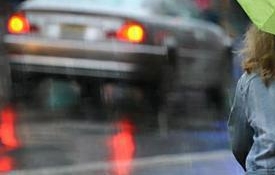}}\vspace{2pt}
 \subfigure[Truth]{\includegraphics[width=.2425\columnwidth]{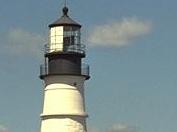}}
  \subfigure[Rain]{\includegraphics[width=.2425\columnwidth]{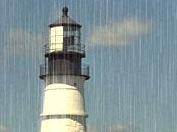}}
  \subfigure[Plain CNN \cite{15}]{\includegraphics[width=.2425\columnwidth]{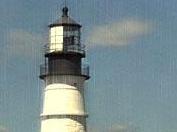}}
 \subfigure[Ours]{\includegraphics[width=.2425\columnwidth]{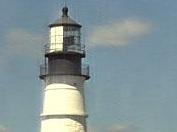}}
 \includegraphics[width=.325\columnwidth]{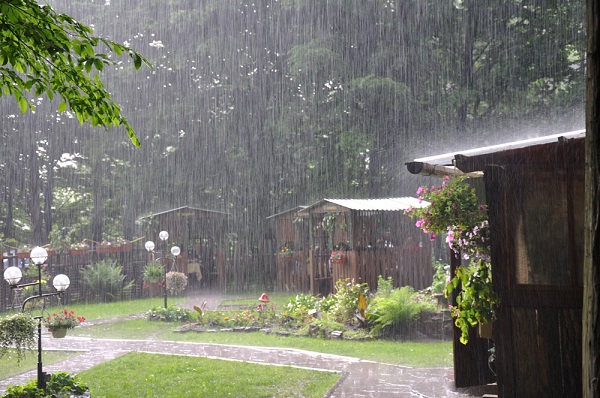}
  \includegraphics[width=.325\columnwidth]{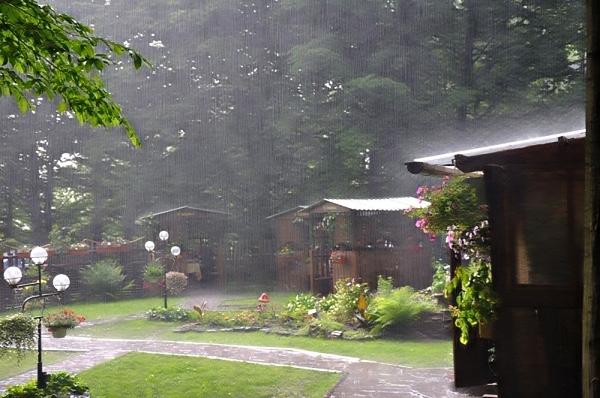}
   \includegraphics[width=.325\columnwidth]{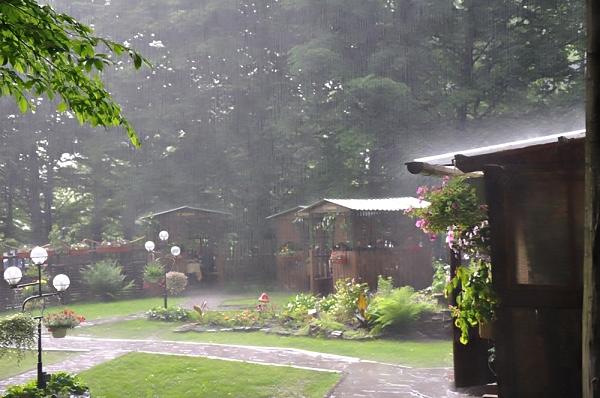}\vspace{2pt}
   \subfigure[Rainy image]{\includegraphics[width=.325\columnwidth]{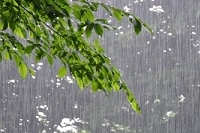}}
  \subfigure[Plain CNN \cite{15}]{\includegraphics[width=.325\columnwidth]{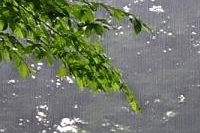}}
   \subfigure[Ours]{\includegraphics[width=.325\columnwidth]{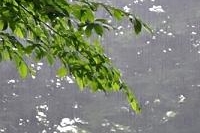}}
   \caption{Top: Two zoomed-in regions of images synthesized with rain. Bottom: Real world rainy image (no enhancement).}\label{fig.CNNvsDerainNet}
\end{figure}

\subsection{Comparison with another potential deep learning method}
The proposed DerainNet combines image domain knowledge as pre-processing before the CNN step. As mentioned \cite{15} proposed directly using a CNN to removing dirt and drops from a window \cite{15}. (This is the only other related deep learning approach we are aware of.) As motivated in Section \ref{sec.Detail_strategy} and Figure \ref{fig.domain}, directly training on the image domain has drawbacks. We show a few other examples on real and synthesized data in Figure \ref{fig.CNNvsDerainNet}. As is evident from these examples as well, directly training on the image domain has drawbacks that are effectively addressed by our approach. We note that both approaches have virtually identical computational complexity.

\subsection{Impact of image enhancement step}
\begin{figure}[h!]
\centering
\subfigure[Rainy image]{
\includegraphics[width = 0.23\textwidth]{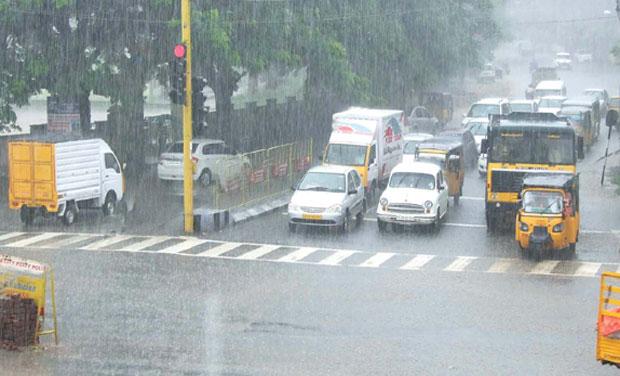}}
\subfigure[De-rain result]{
\includegraphics[width = 0.23\textwidth]{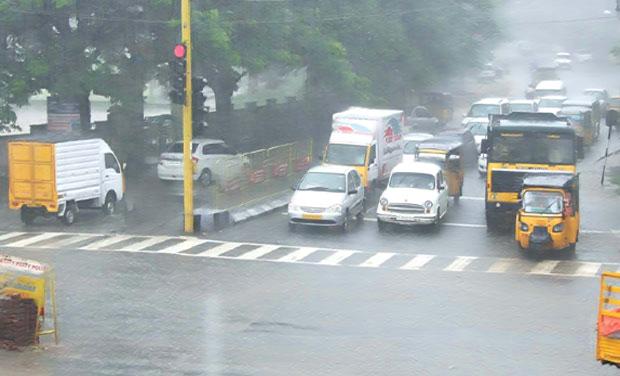}}\\
\subfigure[Post-processed (b)]{
\includegraphics[width = 0.23\textwidth]{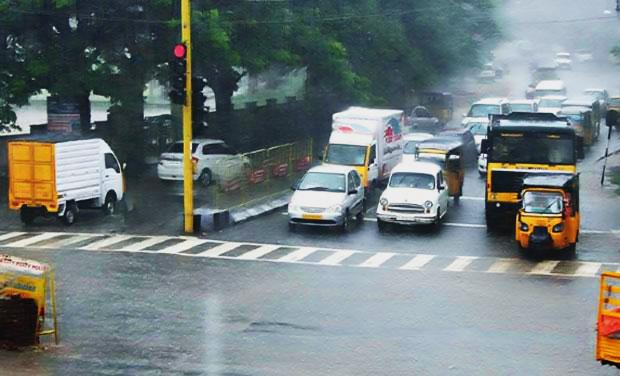}}
\subfigure[Simultaneous processing]{
\includegraphics[width = 0.23\textwidth]{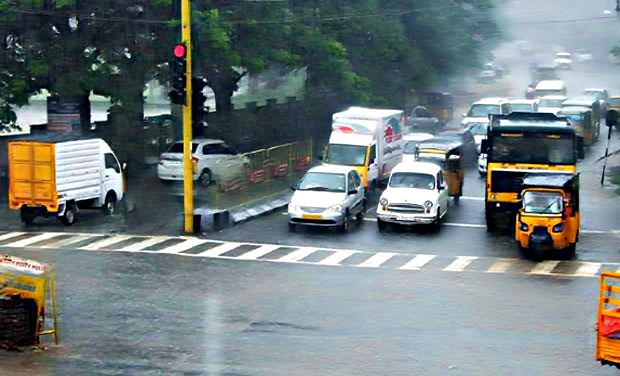}}
\caption{Impact of image enhancement of different processing strategies.} \label{fig.enhance_step}
\end{figure}

In this section we assess the impact of image enhancement on our algorithm. We adopt three processing strategies for real-world data. Specifically, we conduct de-raining without any enhancement, de-raining with the enhancement as a post-processing step after reconstruction, and simultaneous de-raining and enhancement as proposed in Figure \ref{fig.overview}. Figure \ref{fig.enhance_step} shows one example of these different processing strategies. As can be seen, rain streaks are removed by the CNN alone, while the enhancement step further improves the visual quality. We also use the BIQI metric to evaluate the three strategies by testing on collected real-world images, as shown in Table \ref{tab.NIQE1}. Although the visual quality is similar with post-processing, the overall BIQI shows the best quantitative performance of our ``mid-processing'' approach.

\begin{table}[h]
\caption{BIQI results for three enhancement strategies.}
\label{tab.NIQE1}
\centering
\begin{tabular}{|c|c|c|c|c|}
\hline
Images  & No enhance & Post-enhance & Proposed \\
\hline
soccer& 34.69 & \textbf{32.06}    &33.35\\
\hline
Buddha& 37.88&\textbf{29.72}   &28.10\\
\hline
street& 37.47  & 34.33 &\textbf{34.08}\\
\hline
cars & 31.56 &  26.08 &\textbf{24.18}\\
\hline
100 test images& 31.86 $\pm$ 7.89   &  29.98 $\pm$ 7.82  &\textbf{29.86 $\pm$ 6.98}\\
\hline
\end{tabular}
\end{table}

\subsection{Impact of the selected low-pass filter}
\begin{figure}
\centering
\subfigure[Clean image]{\includegraphics[width = .47\columnwidth]{results/examplesdata/2.jpg}}
\subfigure[Rainy image]{\includegraphics[width = .47\columnwidth]{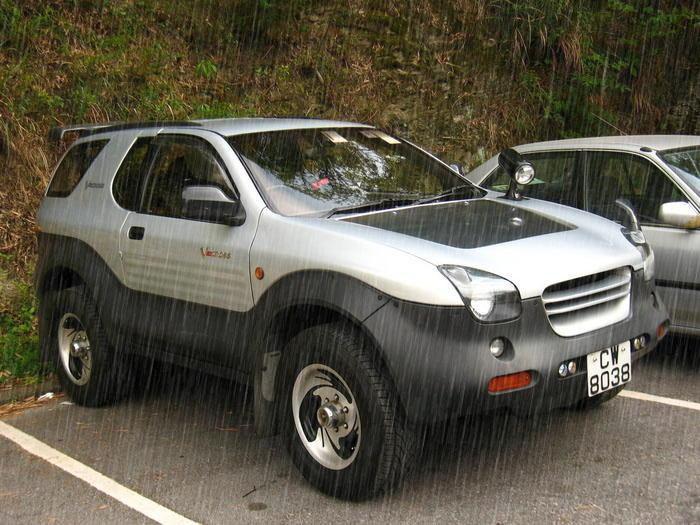}}\\
\subfigure[Base layer \cite{23}]{\includegraphics[width = .32\columnwidth]{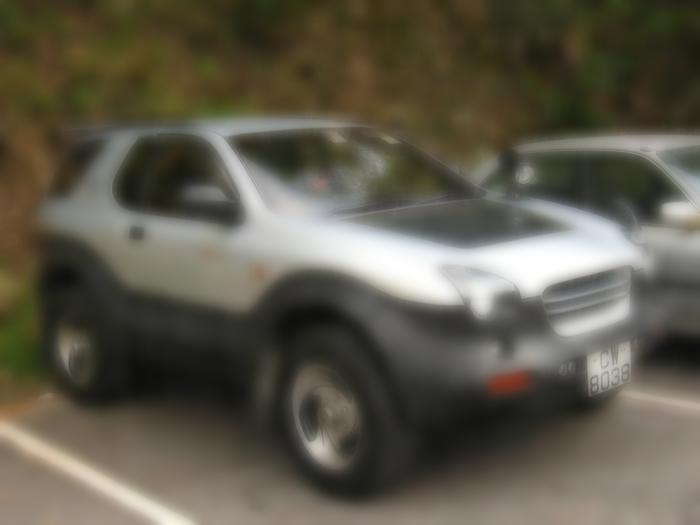}}
\subfigure[Base layer \cite{24}]{\includegraphics[width = .32\columnwidth]{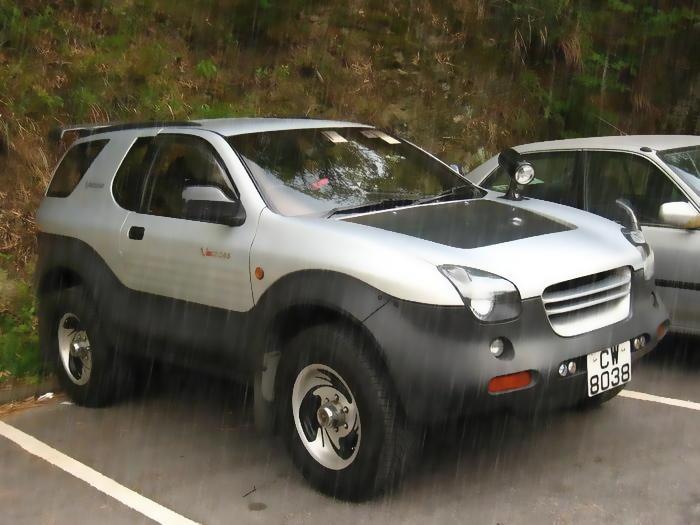}}
\subfigure[Base layer \cite{25}]{\includegraphics[width = .32\columnwidth]{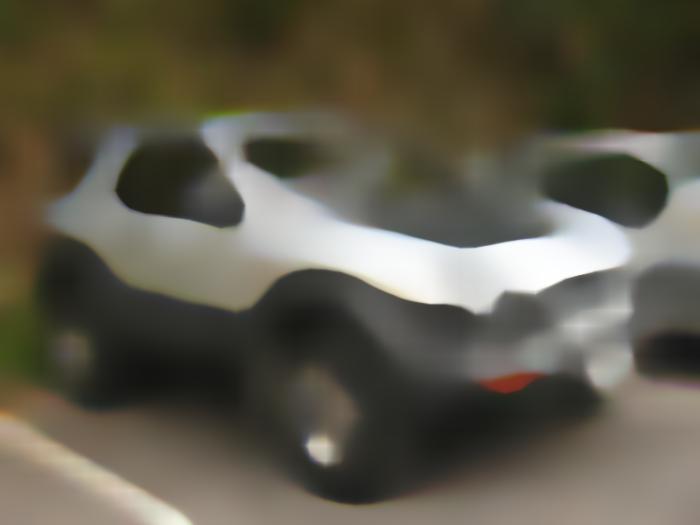}}
\subfigure[Detail layer \cite{23}]{\includegraphics[width = .32\columnwidth]{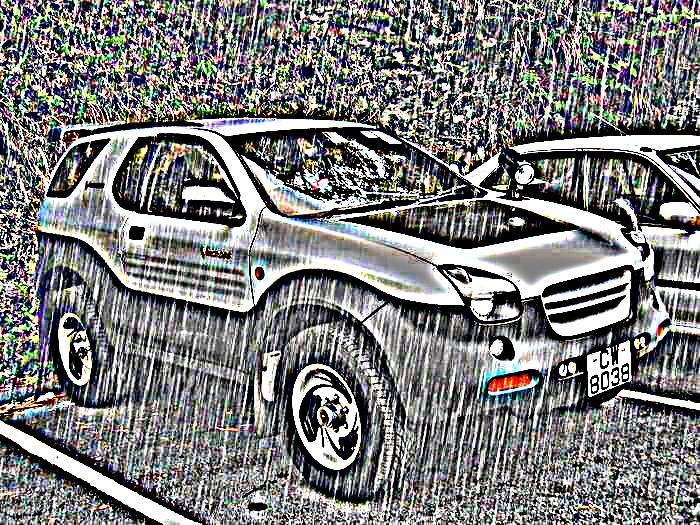}}
\subfigure[Detail layer \cite{24}]{\includegraphics[width = .32\columnwidth]{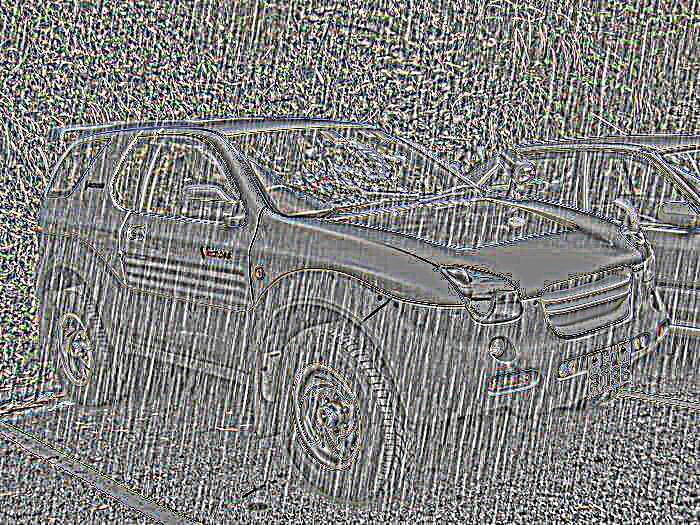}}
\subfigure[Detail layer \cite{25}]{\includegraphics[width = .32\columnwidth]{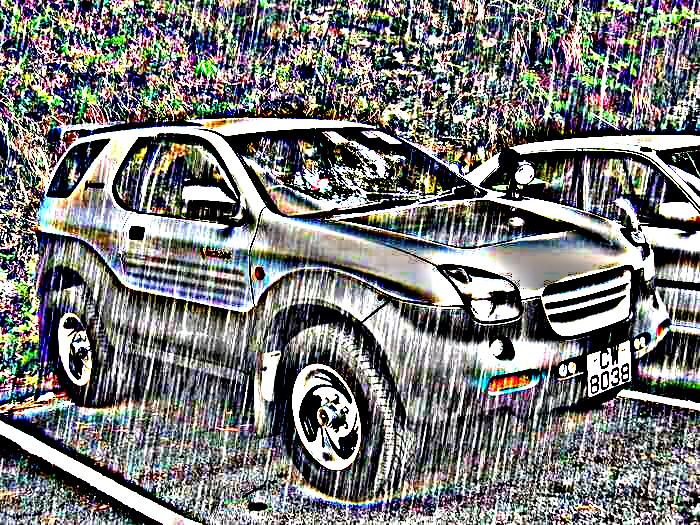}}
\subfigure[De-rained (f)]{\includegraphics[width = .32\columnwidth]{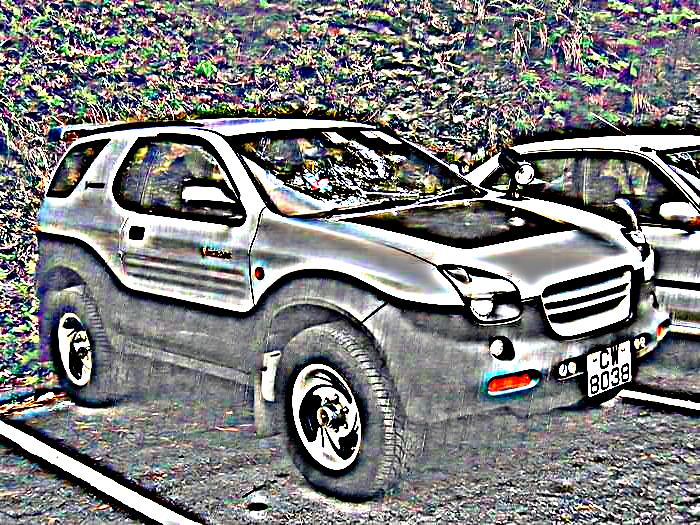}}
\subfigure[De-rained (g)]{\includegraphics[width = .32\columnwidth]{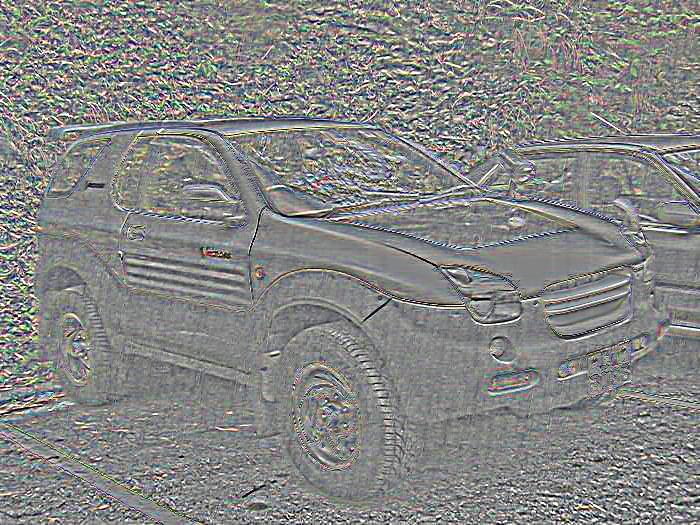}}
\subfigure[De-rained (h)]{\includegraphics[width = .32\columnwidth]{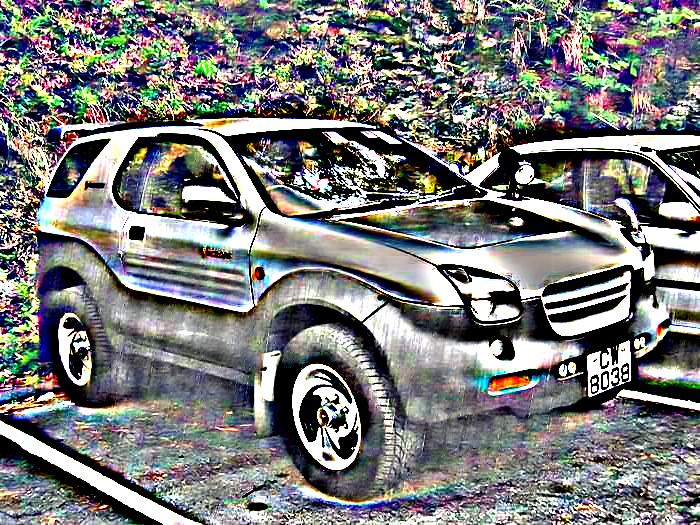}}
\subfigure[Result of (c)$+$(i)]{\includegraphics[width = .32\columnwidth]{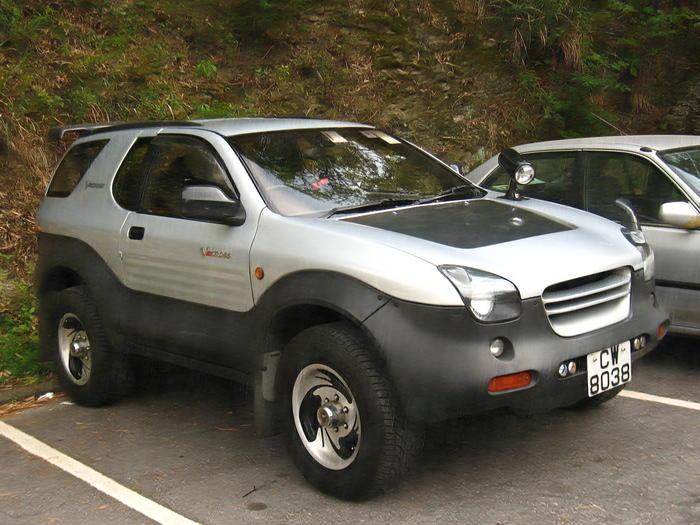}}
\subfigure[Result of (d)$+$(j)]{\includegraphics[width = .32\columnwidth]{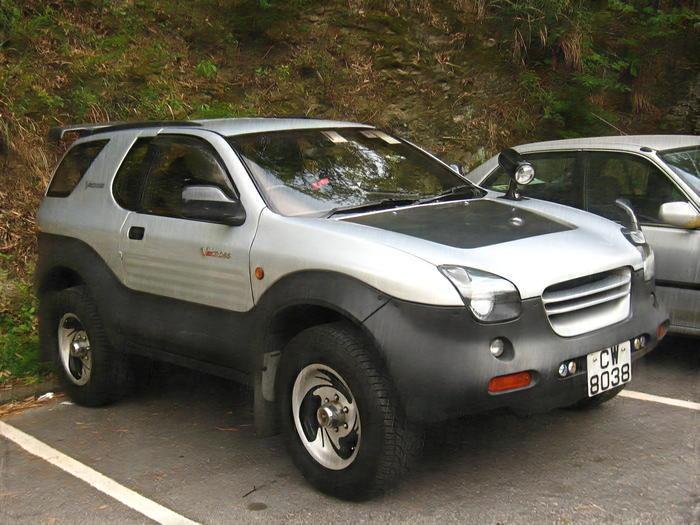}}
\subfigure[Result of (e)$+$(k)]{\includegraphics[width = .32\columnwidth]{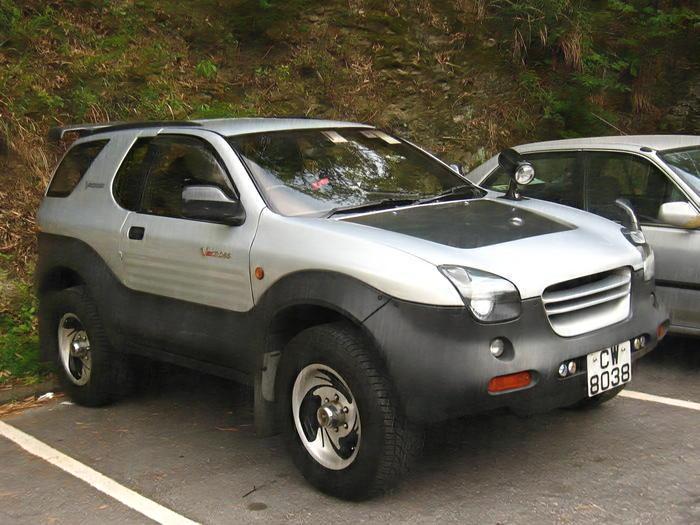}}
\caption{Impact of three low-pass filters \cite{23,24,25}. SSIM values of (f), (j) and (n) are 0.9181, 0.9160 and 0.9158, respectively. Intensities of detail layer images have been amplified for better visualization.} \label{fig.filters}
\end{figure}

Though we choose the guided filter  \cite{23} to separate the base and detail layers for training the CNN, we found that the framework of Figure \ref{fig.overview} is effective using other low-pass filters as well. Figure \ref{fig.filters} shows one example of a de-raining result using different low-pass filters: guided filtering \cite{23}, bilateral filtering \cite{24} and rolling guidance filtering \cite{25}. As can be seen, though the low and high frequency decompositions look significantly different, the de-raining result is qualitatively similar and the three SSIM values of the de-rained results are almost the same. DerainNet is able to recognize and remove rain as long as it is isolated to the detail layer.

\begin{figure*}
\centering
\subfigure[Rainy image]{\includegraphics[width = .18\textwidth]{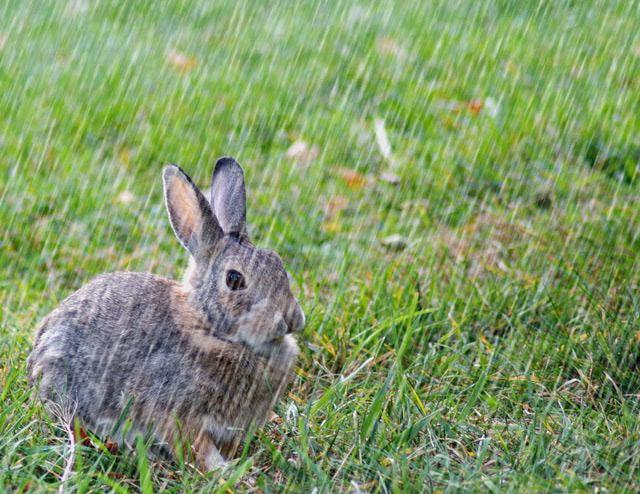}}
\subfigure[Base layer (Y channel)]{\includegraphics[width = .18\textwidth]{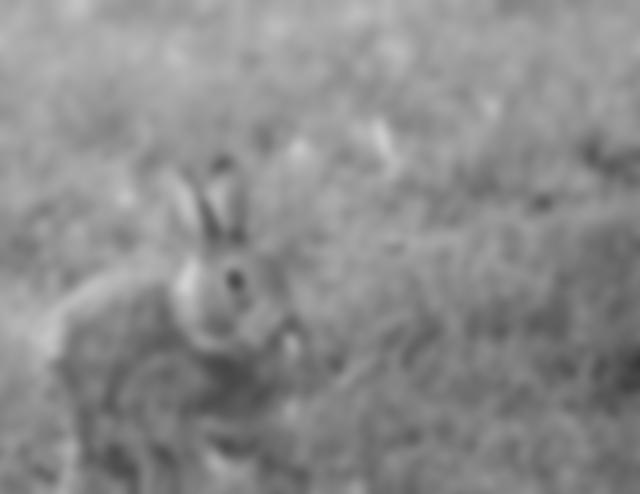}}
\subfigure[Detail layer (Y channel)]{\includegraphics[width = .18\textwidth]{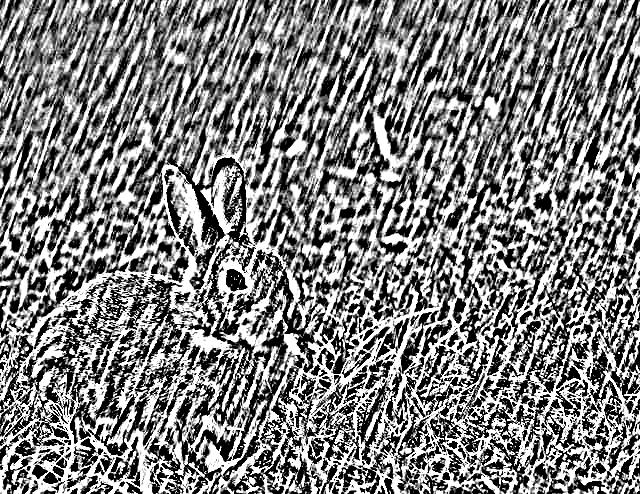}}
\subfigure[De-rained (c) \cite{13}]{\includegraphics[width = .18\textwidth]{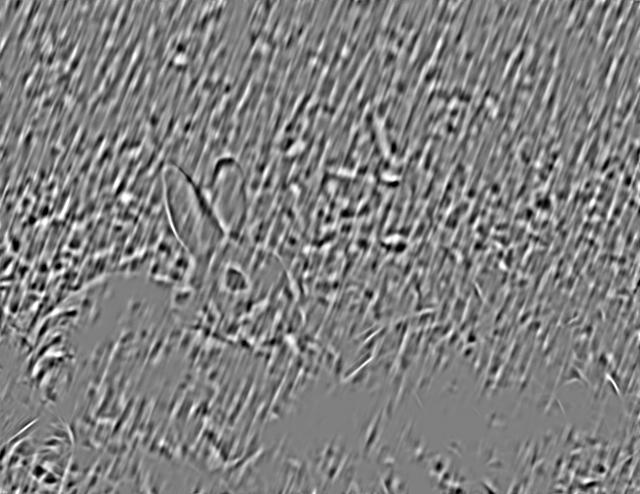}}
\subfigure[De-rained result \cite{13}]{\includegraphics[width = .18\textwidth]{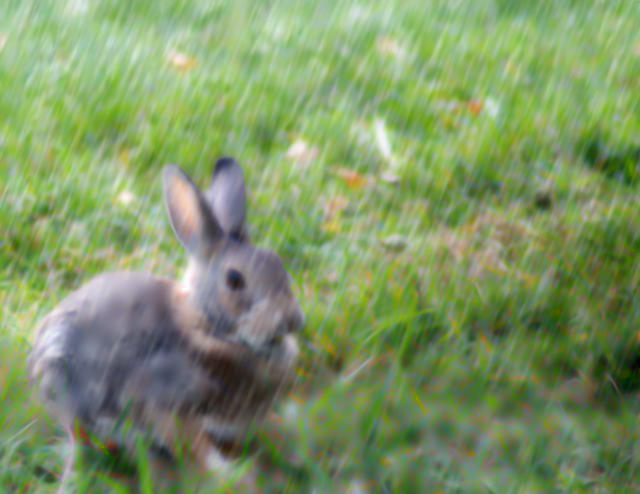}}\\
\subfigure[Clean image]{\includegraphics[width = .18\textwidth]{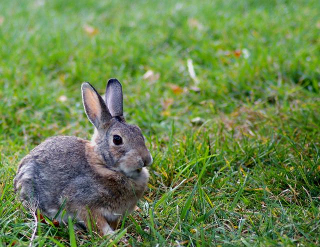}}
\subfigure[Base layer (RGB)]{\includegraphics[width = .18\textwidth]{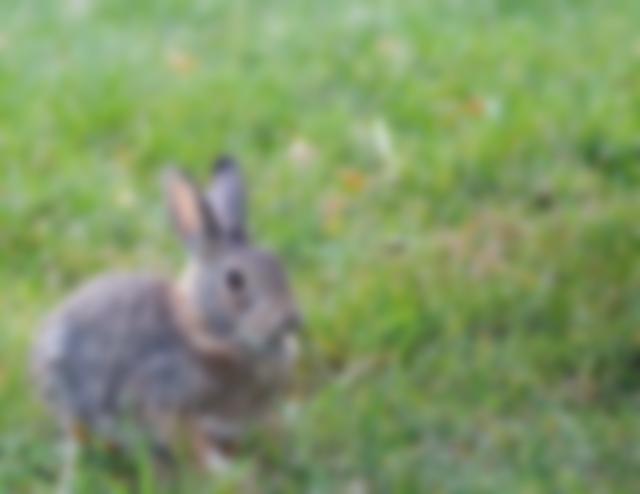}}
\subfigure[Detail layer (RGB)]{\includegraphics[width = .18\textwidth]{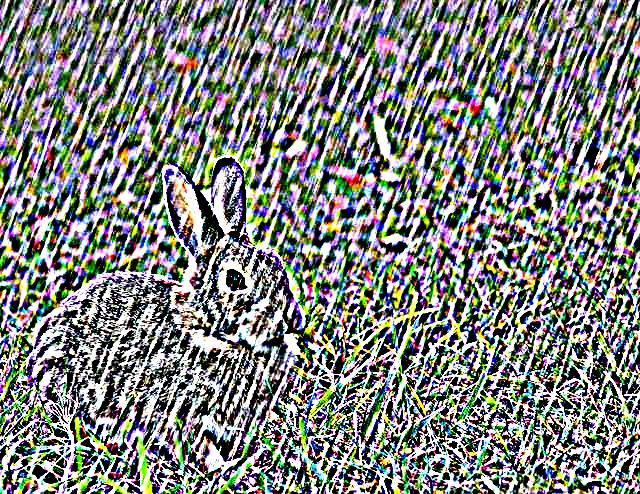}}
\subfigure[Our de-rained (h)]{\includegraphics[width = .18\textwidth]{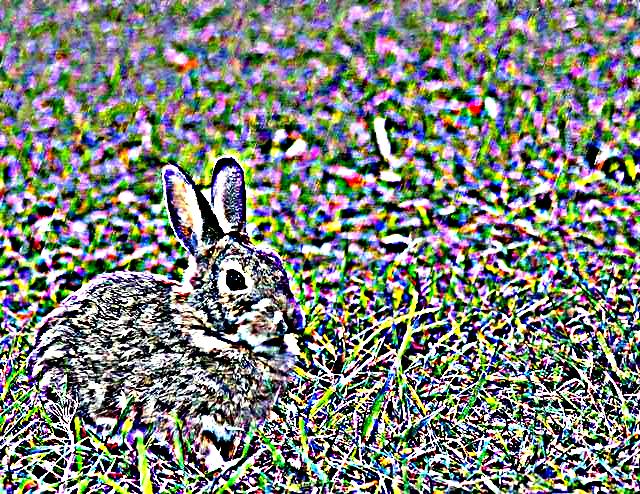}}
\subfigure[Our de-rained result]{\includegraphics[width = .18\textwidth]{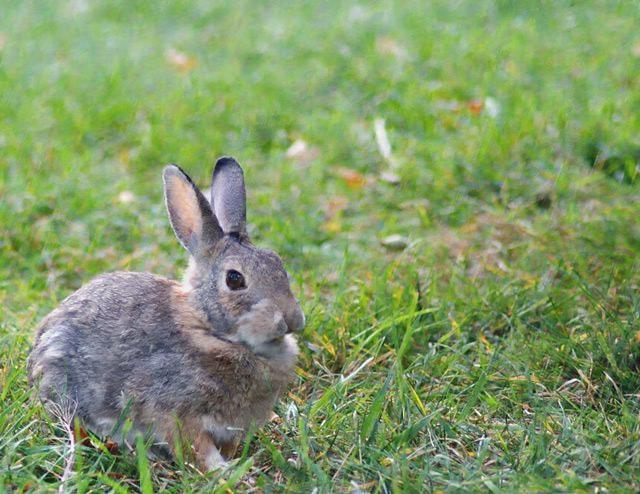}}
\caption{Comparison with method \cite{13} by using the bilateral filtering approach as pre-processing for the respective models. SSIM values of (e) and (j) are 0.77 and 0.87, respectively. Intensities of detail layer images have been amplified for better visualization.} \label{fig.reviewer3}
\end{figure*}

The method proposed in \cite{13} also applies this decomposition strategy using the bilateral filtering, but used in a different model. We make a comparison with method \cite{13} using bilateral filtering for our CNN as well. To ensure the low-pass filter removes all of the rain streaks, we change the default parameters of the bilateral filtering in \cite{13}. Specifically, we change the window size from 5 to 15 and intensity-domain standard deviations from 0.1 to 1. The difference in filtering operations between our method and \cite{13} is that method \cite{13} implements the pre-processing in the Y channel of YUV color space, while our method implements it in the RGB color space. Figure \ref{fig.reviewer3} shows the both intermediate and final de-rained results. As can be seen, both methods isolate the rain to the high-pass portion for de-raining, but \cite{13} fails to completely remove rain streaks in Figure \ref{fig.reviewer3}(c), while our result in Figure \ref{fig.reviewer3}(i) has considerably better 
success. The final results are shown in Figures \ref{fig.reviewer3}(e) and (j).

\begin{figure}
\centering
\includegraphics[width=.9\columnwidth]{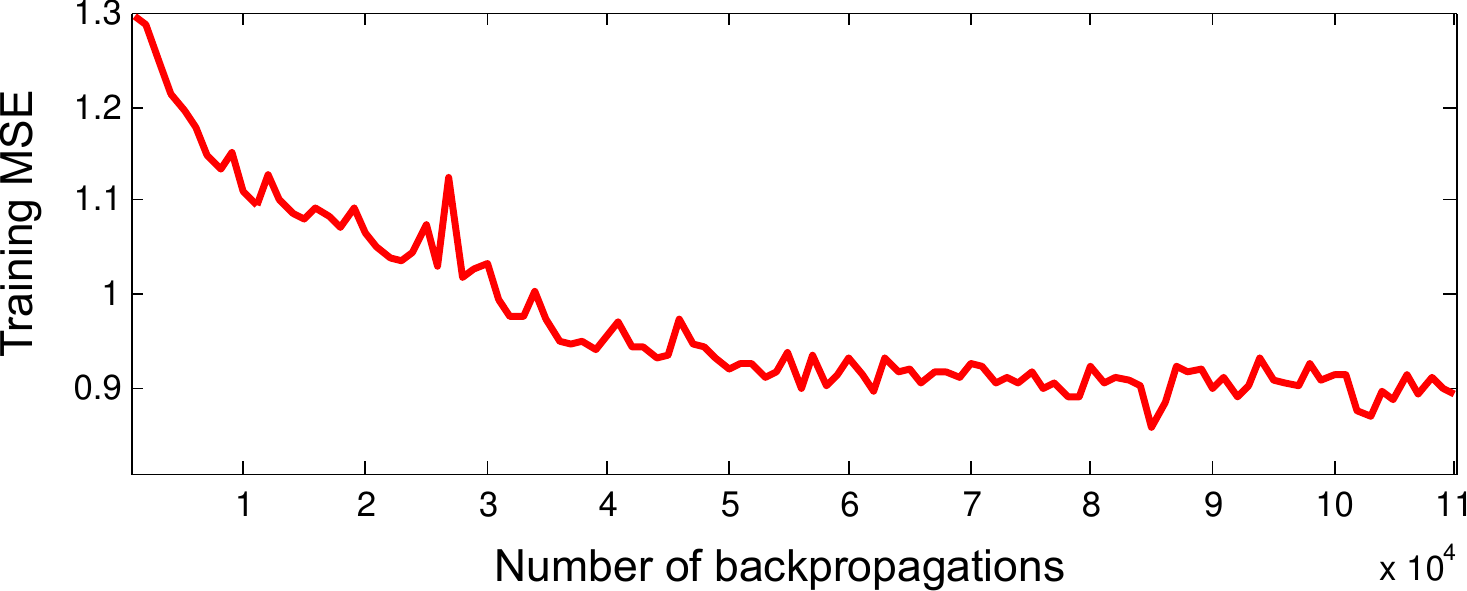}
\caption{The training convergence curve of DerainNet.}\label{fig.convergence}
\end{figure}

\subsection{Training convergence and testing runtime}
Training required approximately two days to run. In Figure \ref{fig.convergence} we show the training convergence as a function of the number of backpropagations.
While training requires a nontrivial amount of computation time, DerainNet is able to process new images very efficiently compared with current state-of-the-art de-raining methods. Table \ref{tab.runtime} shows the average running time for three different image sizes, each averaged over 10 testing images. (Note that these results do not factor in training time, but are for applying these methods to new data.) Since methods \cite{13,16} are based on dictionary learning and method \cite{34} is based on Gaussian mixture model learning, complex optimizations are still required to de-rain new images, leading to a slower computation time. Our method has significantly faster running time since the testing procedure is completely feed-forward after network training. For even larger images, such as those taken by a typical camera, \cite{13,16,34} take from several minutes to over an hour to process a new image, while our method requires roughly half a minute based on a parallel GPU implementation.
\begin{table}
\caption{Comparison of running time (seconds).}
\label{tab.runtime}
\centering
\def\arraystretch{1.25}
\begin{tabular}{|c|c|c|c|c|c|}
\hline
  Image size& Method \cite{13} & Method \cite{16} & Method \cite{34} &Ours\\
\hline
250 $\times$ 250& 68 & 53 & 196  &\textbf{1.3}\\
\hline
500 $\times$ 500 & 76 &  230   &  942  &\textbf{2.8}\\
\hline
750 $\times$ 750   &  99  & 782 &  1374  & \textbf{5.4}\\
\hline
\end{tabular}
\end{table}

\section{Conclusion}
We have presented a deep learning architecture called DerainNet for removing rain from individual images. Using a convolutional neural network on the high frequency detail content, our approach learns the mapping function between clean and rainy image detail layers. Since we do not possess the ground truth clean images corresponding to real-world rainy images, we synthesize clean/rainy image pairs for network learning, and showed how this network still transfers well to real-world images. We showed that deep learning with convolutional neural networks, a technology widely used for high-level vision task, can also be exploited to successfully deal with natural images under bad weather conditions. We also showed that DerainNet noticeably outperforms other state-of-the-art methods with respect to image quality and computational efficiency. In addition, by using image processing domain knowledge, we were able to show that we do not need a very deep (or wide) network to perform this task.

\ifCLASSOPTIONcaptionsoff
  \newpage
\fi

\bibliographystyle{IEEEbib}
\bibliography{derain}
\end{document}